\documentclass[]{style}

\definecolor{lastauthor}{RGB}{143, 68, 115}

\usepackage{marvosym}
\usepackage{eso-pic}

\newcommand{\firstpagelongcatlogo}{%
  \AddToShipoutPictureFG*{%
    \AtPageUpperLeft{%
      \put(\LenToUnit{2.5cm},\LenToUnit{-1.1cm}){%
        \makebox(0,0)[lt]{\includegraphics[height=3.0em]{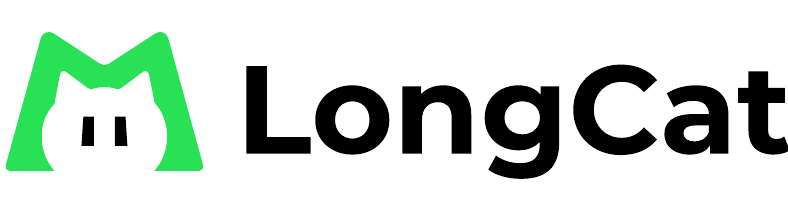}}%
      }%
    }%
  }%
}

\title{dNaViT: Discrete Codes are General-Purpose Visual Representations}

\DeclareRobustCommand{\dnavitteamblock}{%
  \parbox{\linewidth}{\centering
    {\rmfamily\bfseries Meituan LongCat Team}%
  }%
}
\affiliation{\dnavitteamblock}

\usepackage[table]{xcolor}

\usepackage{amsmath}
\usepackage{amssymb}
\usepackage{graphicx}
\usepackage{booktabs}
\usepackage{xcolor}
\usepackage{hyperref}
\usepackage{cleveref}
\usepackage{natbib}
\usepackage{enumitem}
\usepackage{multirow}
\usepackage{subcaption}
\usepackage{microtype}
\usepackage{tabularx}

\usepackage{makecell}

\graphicspath{{resources/figures/}}

\hypersetup{
  colorlinks=true,
  linkcolor=metablue,
  citecolor=metablue,
  urlcolor=metablue,
}

\definecolor{resourcelinkblue}{HTML}{4B8AC9}
\newcommand{\resourcelinkicon}[1]{\makebox[1.25em][c]{\raisebox{-0.15em}{\includegraphics[height=0.9em]{#1}}}\hspace{0.3em}}

\abstract{
Discrete autoregressive modeling, notably through next-token prediction, has become a cornerstone of modern language models. Extending this paradigm to vision allows visual and language models to converge on a unified framework, thereby fostering shared methodology and infrastructure. This work investigates a central question: how far can a finite discrete vocabulary go in representing the richness of the visual world? We present dNaViT, a Discrete Native-resolution Vision Transformer that produces discrete visual primitives analogous to language tokens. Built upon a frozen pretrained encoder, dNaViT applies lightweight residual modulation with a dual-branch objective to inject reconstruction-aware cues into semantic features. By leveraging grouped product quantization, it scales discrete capacity without inflating spatial sequence lengths, while a balanced training recipe maintains both semantic and pixel-level information. Consequently, dNaViT functions as a versatile discrete visual tokenizer, enabling unified encoding across downstream tasks alongside high-fidelity image decoding. Extensive evaluations show that dNaViT matches or exceeds task-specific baselines across 12 visual understanding benchmarks, 4 generation benchmarks, and an image editing benchmark. These findings demonstrate that a compact discrete vocabulary can serve as a universal visual currency and unlock the full potential of fully unified, native multimodal methodology.

\par\vspace{-0.35cm}
\begingroup
\hypersetup{urlcolor=resourcelinkblue}
\noindent\resourcelinkicon{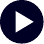}\href{https://lixiaotong97.github.io/dNaViT/}{\textbf{Project Page}}
\hspace{0.6em}%
\resourcelinkicon{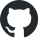}\href{https://github.com/meituan-longcat/LongCat-Next}{\textbf{GitHub}}
\hspace{0.6em}%
\resourcelinkicon{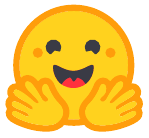}\href{https://huggingface.co/meituan-longcat/LongCat-Next}{\textbf{Hugging Face}}
\endgroup
}

\newcommand{\firstpagecontact}{%
  \begingroup
    \renewcommand{\thefootnote}{}%
    \footnotetext{\textbf{Correspondence:} \email{lixiaotong.edu@gmail.com}}%
  \endgroup
}

\usepackage[dvipsnames]{xcolor}
\definecolor{GainColor}{named}{RawSienna}
\definecolor{BadColor}{named}{Red}

\usepackage{arydshln}
\usepackage[table]{xcolor}  

\newtcolorbox{appendixpromptbox}[1]{
  enhanced,
  breakable,
  colback=metabg!65!white,
  colframe=metablue!72!black,
  colbacktitle=metablue!14!white,
  coltitle=metafg,
  fonttitle=\small\sffamily\bfseries,
  fontupper=\footnotesize\ttfamily,
  title={#1},
  boxrule=0.5pt,
  arc=3pt,
  left=5pt,
  right=5pt,
  top=4pt,
  bottom=4pt,
  before skip=5pt,
  after skip=6pt,
}
\newtcolorbox{appendixrecordbox}[1]{
  enhanced,
  breakable,
  colback=meituanyellow!16!white,
  colframe=meituanlink!82!black,
  colbacktitle=meituanyellow!34!white,
  coltitle=metafg,
  fonttitle=\small\sffamily\bfseries,
  fontupper=\footnotesize,
  title={#1},
  boxrule=0.5pt,
  arc=3pt,
  left=5pt,
  right=5pt,
  top=4pt,
  bottom=4pt,
  before skip=5pt,
  after skip=6pt,
}

\usepackage{epigraph}
\usepackage[normalem]{ulem}
\definecolor{quotemark}{gray}{0.7}
\makeatletter
\def\fquote{%
    \@ifnextchar[{\fquote@i}{\fquote@i[]}%
           }%
\def\fquote@i[#1]{%
    \def\tempa{#1}%
    \@ifnextchar[{\fquote@ii}{\fquote@ii[]}%
                 }%
\def\fquote@ii[#1]{%
    \def\tempb{#1}%
    \@ifnextchar[{\fquote@iii}{\fquote@iii[]}%
                      }%
\def\fquote@iii[#1]{%
    \def\tempc{#1}%
    \vspace{1em}%
    \noindent%
    \begin{list}{}{%
         \setlength{\leftmargin}{0.05\textwidth}%
         \setlength{\rightmargin}{0.05\textwidth}%
                  }%
         \item[]%
         \begin{picture}(0,0)%
         \put(-8,-5){\makebox(0,0){\scalebox{2}{\textcolor{quotemark}{``}}}}%
         \end{picture}%
         \begingroup\itshape}%
 \def\endfquote{%
 \endgroup\par%
 \makebox[0pt][l]{%
 \hspace{0.27\textwidth}%
 \begin{picture}(0,0)(0,0)%
 \put(60,20){\makebox(0,0){%
 \scalebox{2}{\color{quotemark}''}}}%
 \end{picture}}%
 \ifx\tempa\empty%
 \else%
    \ifx\tempc\empty%
       \hfill\rule{110pt}{0.5pt}\\\mbox{}\hfill\tempa,\ \emph{\tempb}%
   \else%
       \hfill\rule{100pt}{0.5pt}\\\mbox{}\hfill\tempa,\ \emph{\tempb},\ \tempc%
   \fi\fi\par%
   \vspace{0.5em}%
 \end{list}%
 }%
 \makeatother

\definecolor{orange}{RGB}{178,92,35}
\definecolor{green1}{RGB}{95,145,51}

\definecolor{red1}{RGB}{197,64,57}
\definecolor{blue1}{RGB}{59,130,220}
\definecolor{green2}{RGB}{82,181,150}
\definecolor{purple1}{RGB}{105,93,223}
\definecolor{orange1}{RGB}{164, 47, 54}
\definecolor{green3}{RGB}{94,145,51}

\definecolor{c_step1}{RGB}{60, 105, 199}
\definecolor{c_step2}{RGB}{159,37,28}
\definecolor{c_step3}{RGB}{107,40,157}

\newenvironment{talign*}
{\csname align*\endcsname}
{\endalign}

\usepackage[utf8]{inputenc}         %
\usepackage[T1]{fontenc}            %
\usepackage{url}                    %
\usepackage{booktabs}               %
\usepackage{amsfonts}               %
\usepackage{nicefrac}               %
\usepackage{microtype}              %
\usepackage{algorithm}
\usepackage{algorithmic}
\usepackage{graphicx}
\usepackage{subcaption}
\usepackage[flushleft]{threeparttable}
\usepackage{float}
\usepackage{multirow}
\usepackage{xspace}
\usepackage{enumitem}
\usepackage[font=small]{caption}
\usepackage{autobreak}
\usepackage{sidecap}
\usepackage{wrapfig}
\usepackage{multicol}
\usepackage{iftex}
\ifXeTeX
  \usepackage{xeCJK}
  \newenvironment{zhtext}{}{}
\else
  \usepackage{CJKutf8}
  \newenvironment{zhtext}{\begin{CJK}{UTF8}{gbsn}}{\end{CJK}}
\fi
\usepackage[toc, page, header]{appendix}
\usepackage{tikz}
\usepackage{pifont}
\usepackage{mdframed}
\usepackage{colortbl}

\definecolor{coral}{RGB}{255,127,80}
\definecolor{darkgreen}{RGB}{0,100,0}
\definecolor{darkyellow}{RGB}{204,153,0}
\definecolor{salmon}{RGB}{250,128,114}
\definecolor{darkred}{RGB}{150,0,0}
\newcommand{\darkredtext}[1]{{\color{darkred}#1}}

\newcommand{\secref}[1]{\hyperref[#1]{\darkredtext{Sec.~\ref*{#1}}}}
\newcommand{\thmref}[1]{\hyperref[#1]{\darkredtext{Thm.~\ref*{#1}}}}
\newcommand{\defref}[1]{\hyperref[#1]{\darkredtext{Def.~\ref*{#1}}}}
\newcommand{\propref}[1]{\hyperref[#1]{\darkredtext{Prop.~\ref*{#1}}}}
\newcommand{\assumpref}[1]{\hyperref[#1]{\darkredtext{Assump.~\ref*{#1}}}}
\newcommand{\remarkref}[1]{\hyperref[#1]{\darkredtext{Rem.~\ref*{#1}}}}
\newcommand{\hypref}[1]{\hyperref[#1]{\darkredtext{Hyp.~\ref*{#1}}}}
\newcommand{\conjref}[1]{\hyperref[#1]{\darkredtext{Conj.~\ref*{#1}}}}
\newcommand{\lemref}[1]{\hyperref[#1]{\darkredtext{Lem.~\ref*{#1}}}}
\newcommand{\corref}[1]{\hyperref[#1]{\darkredtext{Cor.~\ref*{#1}}}}
\newcommand{\noteref}[1]{\hyperref[#1]{\darkredtext{Nota.~\ref*{#1}}}}
\newcommand{\claimref}[1]{\hyperref[#1]{\darkredtext{Clm.~\ref*{#1}}}}
\newcommand{\obsref}[1]{\hyperref[#1]{\darkredtext{Obs.~\ref*{#1}}}}
\newcommand{\algref}[1]{\hyperref[#1]{\darkredtext{Alg.~\ref*{#1}}}}
\newcommand{\figref}[1]{\hyperref[#1]{\darkredtext{Figure~\ref*{#1}}}}
\newcommand{\tabref}[1]{\hyperref[#1]{\darkredtext{Table~\ref*{#1}}}}
\newcommand{\appref}[1]{\hyperref[#1]{\darkredtext{App.~\ref*{#1}}}}
\renewcommand{\eqref}[1]{\hyperref[#1]{\darkredtext{Eq.~\ref*{#1}}}}

\usepackage{listings}
\usepackage{xcolor}
\usepackage{booktabs}
\usepackage{amssymb}      %
\usepackage{tikz}         %

\lstdefinestyle{agentloop}{
  basicstyle=\ttfamily\scriptsize,
  commentstyle=\color{gray!70!black}\itshape,
  keywordstyle=\color{blue!55!black}\bfseries,
  showstringspaces=false,
  columns=fullflexible,
  keepspaces=true,
  escapeinside={(*@}{@*)},
  morekeywords={while,if,not,is,continue,and,or,True,False,None,def,return},
  frame=tb,
  framerule=0.5pt,
  rulecolor=\color{gray!60},
  framesep=5pt,
  backgroundcolor=\color{gray!4},
  xleftmargin=8pt,
  xrightmargin=4pt,
  aboveskip=2pt,
  belowskip=2pt,
  breaklines=false,
}

\definecolor{darkred}{RGB}{165,42,42}

\makeatletter
\newcommand{\dnavitrowhighlight}{%
  \leavevmode\rlap{%
    \color[HTML]{ECEBFF}%
    \rule[-\dp\@arstrutbox]{\linewidth}{%
      \dimexpr\ht\@arstrutbox+\dp\@arstrutbox\relax}%
  }\ignorespaces
}
\makeatother

\begin{document}
\maketitle
\firstpagelongcatlogo
\firstpagecontact
\begin{figure}[H]
  \centering
  \includegraphics[width=\textwidth]{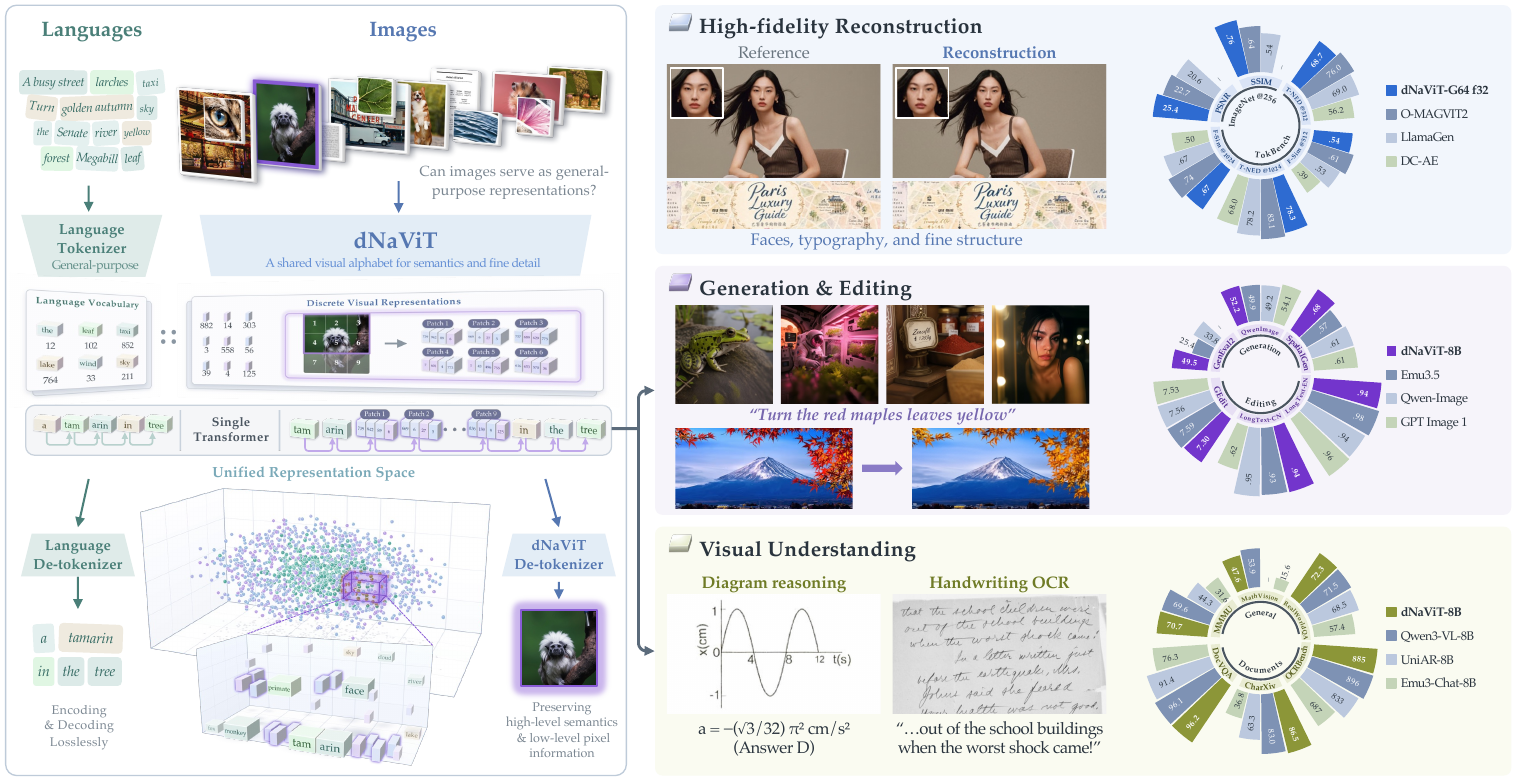}
  \caption{\textbf{dNaViT: A unified visual tokenizer for general-purpose modeling.}
dNaViT represents arbitrary-resolution images as discrete visual primitives, natively unifying \textit{reconstruction, understanding, generation, and editing} under a shared vocabulary. By achieving performance competitive with specialized baselines across all domains, dNaViT validates a shared token-based paradigm for vision, analogous to language modeling.}
  \label{fig:autodesign-poster-showcase}
\end{figure}
\clearpage
\begin{figure}[p]
  \centering
  \includegraphics[width=\textwidth,height=0.93\textheight,keepaspectratio]{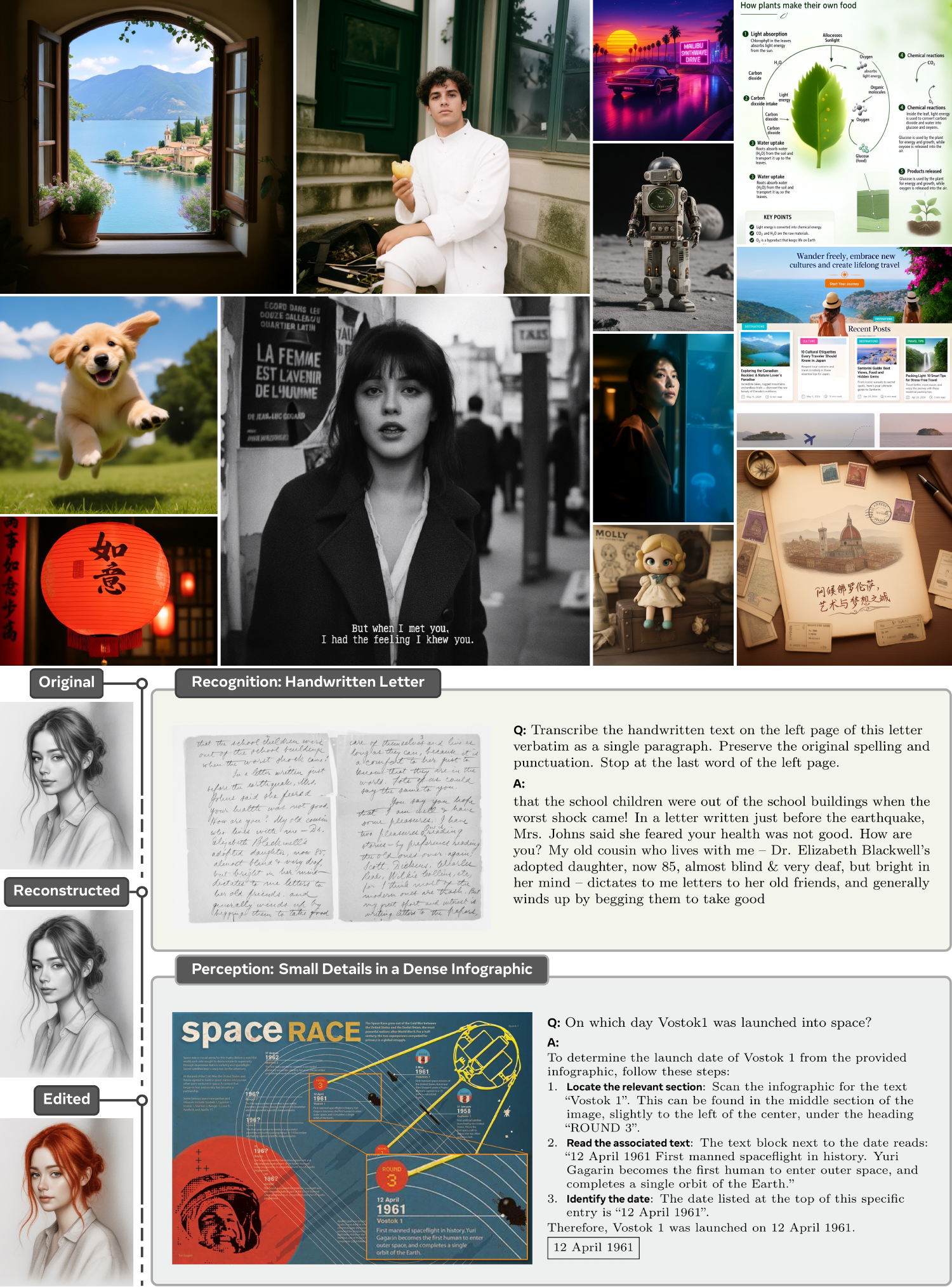}
  \caption{\textbf{dNaViT's unified discrete representation as a foundation for diverse downstream visual tasks.} Qualitative showcases illustrate the versatility of this shared representation.}
  \label{fig:downstream_showcases}
\end{figure}
\clearpage
\setcounter{tocdepth}{2}
\tableofcontents
\clearpage

\section{Introduction}
\label{sec:intro}

Language models have achieved remarkable success through next-token prediction, anchored by discrete tokens as their fundamental modeling unit~\citep{brown2020fewshot}. From the perspective of \emph{compression as intelligence}~\citep{deletang2023compression}, discrete tokens provide a unified substrate for representation learning and autoregressive modeling. In contrast, modern multimodal models frequently rely on modality-specific representations and heterogeneous learning objectives~\citep{caffagni2024revolution}. This paradigm divergence disconnects vision from the established techniques and scalable infrastructure to language, raising a foundational question: can visual information be represented losslessly like language using a learned discrete vocabulary?

Discrete visual representations offer a promising bridge, naturally aligning visual inputs with autoregressive language architectures under a unified token-prediction objective~\citep{xie2024showo,wang2024emu3,zou2026omnimamba,peng2026unified,longcatnext2026,cui2025emu35}. However, discretizing dense, continuous visual signals into a finite vocabulary presents a severe information bottleneck~\citep{ma2025unitok,yu2026viq} that hinders performance. Crucially, harmonizing high-level semantic abstraction for understanding with fine-grained pixel details for high-fidelity reconstruction within a single discrete bottleneck remains an unresolved tension~\citep{wu2024janus,qu2025tokenflow}.

To answer the question posed above and address these challenges, we introduce \textbf{dNaViT}, a Discrete Native-resolution Vision Transformer that extends our preliminary work in LongCat-Next~\citep{longcatnext2026}. Following the representation encoder paradigm~\citep{zheng2025rae}, dNaViT enriches the semantic features of a frozen pretrained encoder with pixel details through a \textit{Residual Modulator} (ResMod), without compromising its underlying semantic capability. The resulting features are discretized into a shared vocabulary using Grouped Product Quantization (PQ)~\citep{jegou2011pq} under dual-branch representation and pixel reconstruction loss. Training progresses through continuous initialization, discrete feature alignment, and final joint optimization, which balance theses objectives and suppress gradient interference.

These design choices yield a shared discrete interface for tasks that conventionally rely on separate visual encodings, extending the utility of pretrained representations beyond their original objectives. By rotating and translating pretrained features, residual modulation reorganizes their geometry to accommodate reconstruction-relevant information, making more effective use of the available representational capacity. Combined with sufficient factorized code capacity, this allows semantic structure and fine-grained detail to coexist within a shared discrete bottleneck, mitigating information loss during quantization. The resulting tokenizer brings \textit{the flexibility of text tokenization} to vision: images can be encoded into discrete sequences and decoded into faithful reconstructions, while the codes retain semantic information closely aligned with language.

We evaluate the generality of dNaViT across four core capabilities: image reconstruction, classification, multimodal understanding, generation, editing. Driven by a single shared discrete vocabulary with a spatial downsampling factor $f=32$, dNaViT achieves $28.12\text{~dB}$ PSNR and $0.81$ SSIM at $512\times512$ resolution without sacrificing semantic representation capability. When paired with Qwen3-8B, dNaViT surpasses prior discrete baselines and rivals Qwen3-VL-8B across key document and reasoning benchmarks, scoring $885$ on OCRBench. For image generation and editing, dNaViT achieve impressive performance on GenEval2 ($49.5$) and outperforms open-weight baselines on Qwen-Image-Bench ($52.2$), while excelling at bilingual text rendering. Crucially, dNaViT establishes these comprehensive understanding and generation capabilities under a substantially reduced data budget relative to existing models.

These findings suggest that the gap between continuous visual signals and discrete tokens is a challenge of representation learning rather than an inherent limitation. With appropriate modeling, discrete codes can serve as general-purpose visual primitives, bringing vision and language into a shared token-based paradigm. Our main contributions are summarized as follows:

\begin{enumerate}[leftmargin=*,noitemsep]
  \item This work explores the question of \emph{whether discrete codes can serve as general-purpose visual primitives}, analogous to the role of tokens in language modeling. To investigate this possibility, we introduce dNaViT, a native-resolution visual tokenizer designed to capture both high-level semantic structure and fine-grained pixel information within a shared vocabulary for different tasks.
  \item dNaViT uses residual modulation to aggregate multi-level features from a frozen vision encoder and grouped product quantization to encode the enriched representations. Progressive training coordinates dual-branch semantic and pixel reconstruction objectives. Experiments demonstrate that this design balances semantic preservation and reconstruction fidelity within the discrete bottleneck.
  \item Comprehensive evaluations across visual understanding, image reconstruction, generation, and editing demonstrate dNaViT's generality. With a substantially smaller training data budget than the compared models, dNaViT achieves strong understanding and generation performance, highlighting the effectiveness and data efficiency of its unified, language-aligned representation.
  \item This work demonstrates that a shared discrete token space can serve as a \emph{universal visual currency}, providing a general-purpose intermediate representation across complementary tasks and a foundation for fully unified, native multimodal modeling.
\end{enumerate}

\begin{figure}[!t]
  \centering
  \includegraphics[width=\textwidth]{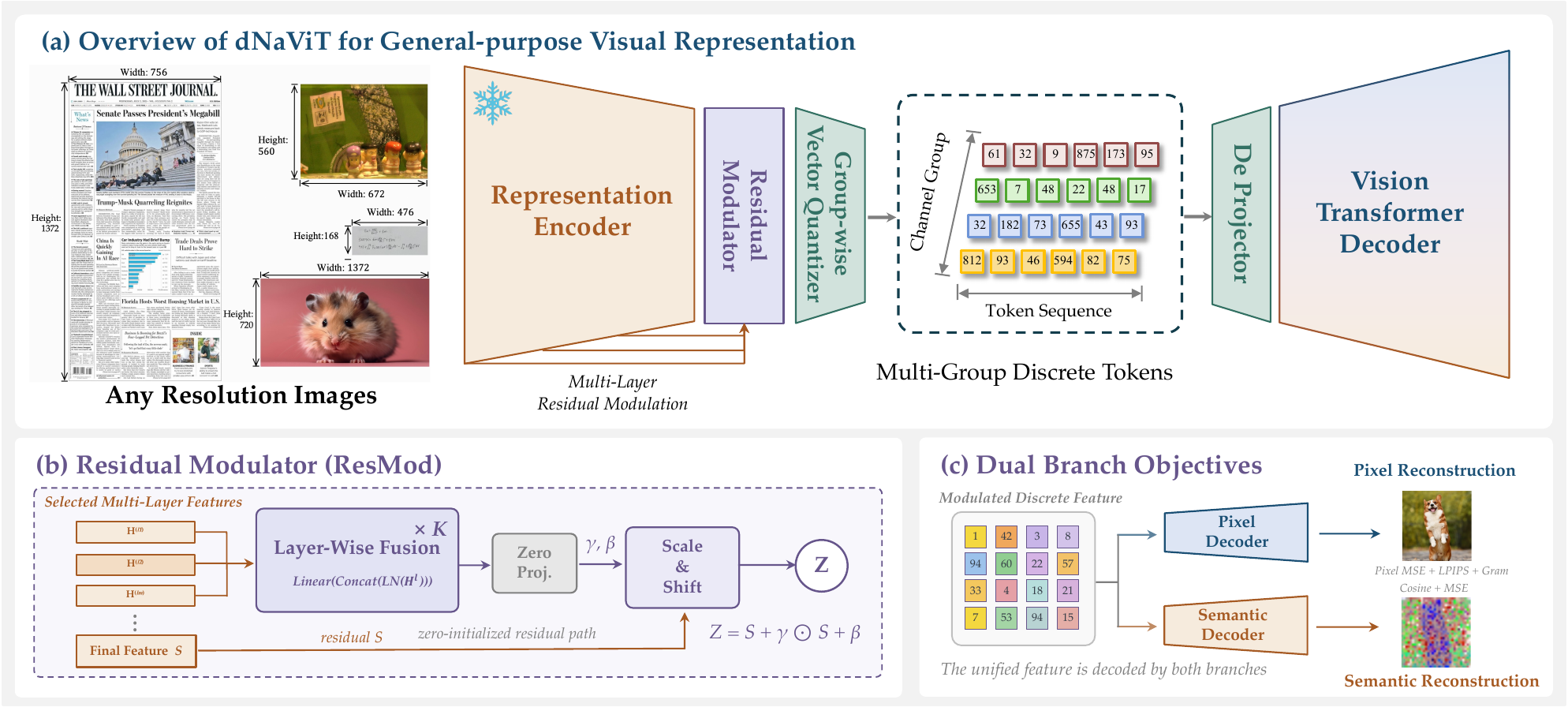}
  \caption{\textbf{Overview of dNaViT.}
  (a) A frozen representation encoder handles arbitrary resolutions,
  with ResMod injecting multi-level information before group-wise product
  quantization. The shared discrete codes are decoded by individual ViT decoders into pixel and semantic reconstructions.
  (b) ResMod fuses intermediate features and predicts a zero-initialized
  scale-and-shift residual.
  (c) Pixel and semantic branches are optimized with image and feature
  reconstruction losses, respectively.}
  \label{fig:dnavit_overview}
\end{figure}

\section{dNaViT}

This section presents dNaViT, a visual tokenizer that learns shared discrete codes as a unified interface for diverse downstream tasks, replacing separate task-specific feature representations. We will describe how the architecture supports these capabilities while mitigating conflicts between semantic abstraction and fine-grained detail preservation, as well as information loss during discretization.

\subsection{Background}
\label{sec:background}

Visual modeling has largely relied on distinct continuous representations for different downstream tasks: pretrained semantic features for multimodal understanding~\citep{siglip2} and reconstruction-oriented latents for diffusion-based synthesis~\citep{ldm}. Representation autoencoders bridge these approaches by pairing pretrained semantic encoders with reconstruction decoders~\citep{zheng2025rae}. By contrast, language models use discrete tokens as a uniform representation of text, supporting a unified, scalable next-token prediction framework. Meanwhile, recent benchmarks~\citep{an2026genius,yan2026worldcast} highlight the growing need to integrate linguistic and visual intelligence to solve real-world tasks.

Extending this framework to vision is challenging: unlike the discrete symbols of language, visual signals encode dense, continuous variation. Discrete visual tokenization addresses this by mapping images to sequences of codebook indices. For example, vector-quantized autoencoders assign continuous features to learned codebook entries and reconstruct images from the corresponding embeddings~\citep{van2017neural}. These indices can be embedded and predicted like text tokens, enabling a shared autoregressive interface and the reuse of language-model architectures and objectives. Chameleon~\citep{chameleon2024mixed} and Emu3~\citep{wang2024emu3} demonstrate this formulation for multimodal understanding and generation. However, information loss during tokenization can constrain downstream performance and the benefits of scaling.

A shared token interface alone does not guarantee that the learned codes jointly preserve semantic information and fine-grained visual detail, both of which are essential to general-purpose visual modeling. Prior studies have examined the distinct information requirements of understanding and reconstruction~\citep{qu2025tokenflow} and identified limited discrete capacity as a key bottleneck in supporting both objectives~\citep{ma2025unitok}. Motivated by these findings, dNaViT aims to preserve complementary task-relevant information within a shared discrete vocabulary and match the performance of task-specific representations.

\subsection{Overall Architecture}
Motivated by the exploration of information recovery through residual connections~\citep{longcatnext2026}, dNaViT further propose residual modulation to aggregate complementary cues and enrich the features of a frozen native-resolution vision encoder with reconstruction-relevant detail (Fig.~\ref{fig:dnavit_overview}). To preserve both semantic structure and visual detail within a shared discrete bottleneck, we quantize the enriched features using grouped product quantization and supervise the resulting codes through dual reconstruction branches. Progressive training coordinates these objectives to mitigate interference during optimization.

\subsection{Representation Encoder}
\label{sec:representation_encoder}
Unified visual tokenization must reconcile appearance invariance for understanding with sensitivity to detail for reconstruction. RAEs demonstrate reconstruction from continuous semantic features~\citep{zheng2025rae}, while LongCat-Next suggests that residual connections can carry low-level information across a semantic encoder~\citep{longcatnext2026}. This motivates us to directly leverage complementary cues by recombining multiple encoder depths, without even training the encoder itself. Meanwhile, quantization may collapse weakly expressed appearance differences into the same code. It therefore requires retaining the final semantic representation as an explicit anchor throughout quantization to preserve representation quality.

To tackle this, we introduce a
\emph{Residual Modulator} (ResMod) inspired by frozen semantic base with
detail-aware modulation of PAE~\citep{yue2026pae}, which fuses intermediate features from
the frozen encoder and learns a spatially varying correction to its final
representation. Multi-level fusion exposes complementary spatial cues, and a
zero-initialized residual update preserves the pretrained features at
initialization. The resulting feature grid provides the quantizer with a
shared representation whose semantic structure and local detail can be
supervised by the two reconstruction branches.

\paragraph{\textbf{\upshape Residual Modulator.}}
Figure~\ref{fig:dnavit_overview}(b) illustrates the module. Let
$\{\mathbf{H}^{(\ell)}\}_{\ell=0}^{L}$ be the spatial tokens of the frozen
vision transformer, with
$\mathbf{H}^{(\ell)}\in\mathbb{R}^{N\times D}$. We select early and deeper
layers $\mathcal{S}\subseteq\{0,\ldots,L\}$ and fuse their normalized features:
\begin{equation}
\mathbf{U}
=\mathcal{T}_{\theta}\!\left(
\mathcal{F}_{\theta}\!\left(
\operatorname{Concat}_{\ell\in\mathcal{S}}
\left[\operatorname{LN}_{\ell}(\mathbf{H}^{(\ell)})\right]
\right)\right),
\label{eq:resmod_fusion}
\end{equation}
where $\mathcal{F}_{\theta}$ projects the concatenated features to a common
width and $\mathcal{T}_{\theta}$ is a lightweight stack of self-attention
and feed-forward blocks. These intermediate taps provide direct access to
spatial information at different levels of abstraction.

Let $\mathbf{S}=\mathcal{A}(\mathbf{H}^{(L)})$ denote the semantic base on
the quantizer grid, where $\mathcal{A}$ is the identity or the encoder's
spatial aggregation operator. We write $N_q$ and $C$ for the number of
spatial positions and the channel width on this grid, respectively.
A projection $\mathcal{P}_{\theta}$ aligns
$\mathbf{U}$ to this grid, and an affine head predicts token-wise scale
and shift parameters:
\begin{equation}
(\boldsymbol{\gamma},\boldsymbol{\beta})
=\operatorname{Split}\!\left(
\mathcal{W}_{\theta}
\bigl(\operatorname{LN}(\mathcal{P}_{\theta}(\mathbf{U}))\bigr)
\right),\qquad
\mathbf{Z}
=\mathbf{S}
+\underbrace{\boldsymbol{\gamma}\odot\mathbf{S}
+\boldsymbol{\beta}}_{\boldsymbol{\Delta}_{\theta}}.
\label{eq:resmod_modulation}
\end{equation}
The projection $\mathcal{P}_{\theta}$ also aggregates spatial tokens when
$\mathbf{S}$ uses a coarser grid. We initialize
$\mathcal{W}_{\theta}$ to zero, giving $\boldsymbol{\Delta}_{\theta}=\mathbf{0}$ and
$\mathbf{Z}=\mathbf{S}$ at initialization. Training then learns how to
reweight the semantic base and add complementary information at each
position. Thus, residual pathways within the encoder provide access to
information across depth, and ResMod organizes this information for the
discrete bottleneck while keeping the pretrained encoder fixed. As a result, ResMod only translates and rotates the pretrained features to enlarge representation capacity while preserving their original semantic structure.

\begin{figure}[t]
  \centering
  \includegraphics[width=\textwidth]{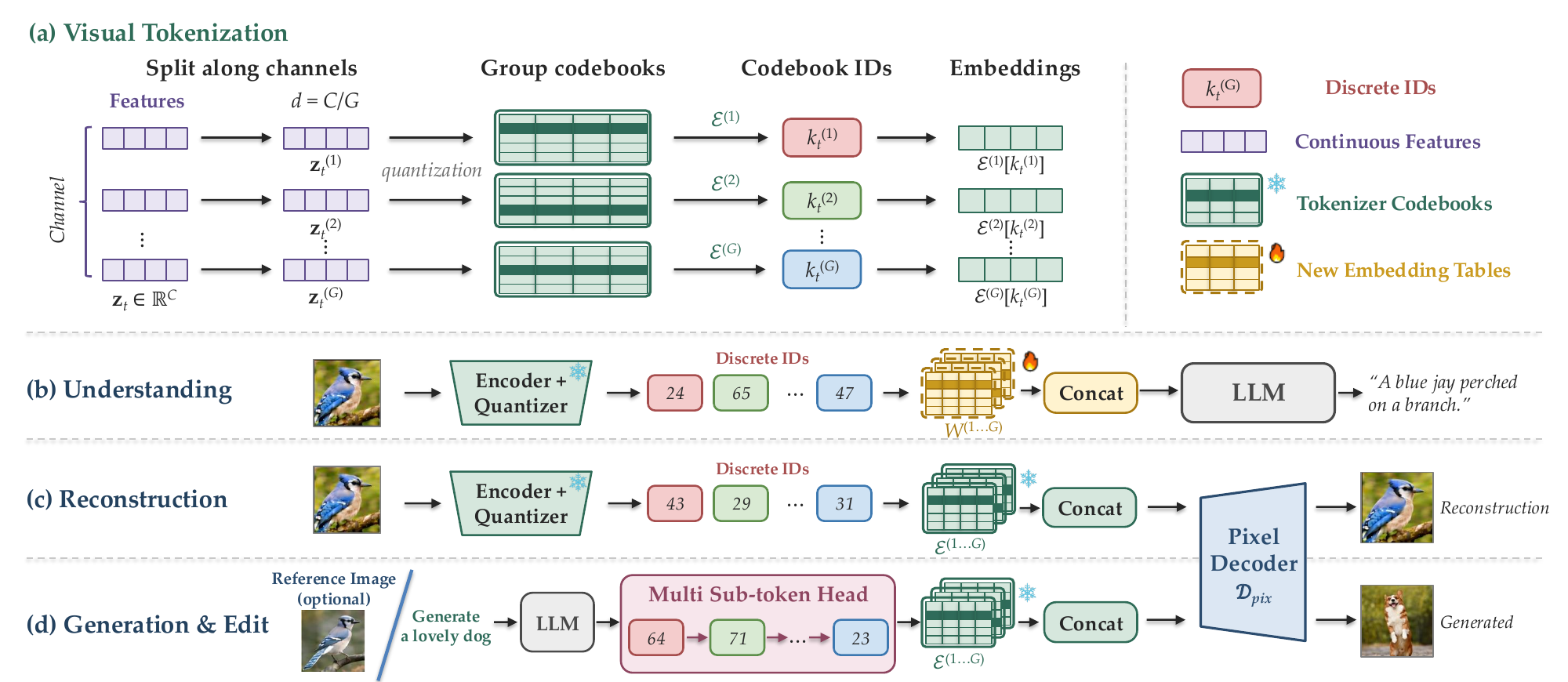}
  \caption{\textbf{Tokenization and Unified Codes for Downstream Tasks.}
  (a) At spatial position $t$, the modulated feature $\mathbf{z}_t$ is split into $G$ channel groups and quantized with group-specific codebooks $\mathcal{E}^{(g)}$, yielding indices $(k_t^{(1)},\ldots,k_t^{(G)})$ and their embeddings.
(b) Understanding supplies only code indices to the language model (LM) through newly initialized group-wise embedding tables $\{\mathbf{W}^{(g)}\}_{g=1}^{G}$, trained jointly with the LM.
(c) Reconstruction concatenates the indexed embeddings from the frozen tokenizer codebooks and feeds them to the pixel decoder $\mathcal{D}_{\mathrm{pix}}$.
(d) Generation and editing use a multi-sub-token head atop the LM to predict each position's multi-level indices, optionally conditioned on a reference image, then decode through the same frozen codebooks and pixel decoder.}
  \label{fig:quantizer}
\end{figure}

\subsection{Quantizer}\label{sec:quantizer}

The residual modulator enriches the features entering the discrete bottleneck, but retaining the exposed semantic distinctions and spatial detail requires sufficient quantization capacity at each spatial position. Motivated by evidence that multi-codebook quantization with a sufficient bit budget preserves semantic information at a level competitive with continuous representations~\citep{zou2026can}, we adopt grouped product quantization~\citep{jegou2011pq} to expand this capacity while keeping the spatial token grid unchanged. As illustrated in Fig.~\ref{fig:quantizer}, the features are factorized along the channel dimension and encoded using multiple codebooks, avoiding a prohibitively large monolithic vocabulary.

Let $\mathbf{z}_t\in\mathbb{R}^{C}$ denote the feature presented to the quantizer at position $t$ of the modulated representation $\mathbf{Z}\in\mathbb{R}^{N_q\times C}$. We partition its channels into $G$ disjoint groups, $\mathbf{z}_t=\operatorname{Concat}_{g=1}^{G}\left[\mathbf{z}_t^{(g)}\right]$, where each group has dimension $d=C/G$. Each group is assigned its own learned codebook $\mathcal{E}^{(g)}=\{\mathbf{e}_k^{(g)}\}_{k=1}^{K}$, with $\mathbf{e}_k^{(g)}\in\mathbb{R}^{d}$. Quantization selects the nearest entry within each group and concatenates the selected embeddings:
\begin{equation}
\begin{gathered}
k_t^{(g)}
=\operatorname*{arg\,min}_{k\in\{1,\ldots,K\}}
\left\|\mathbf{z}_t^{(g)}-\mathbf{e}_k^{(g)}\right\|_2^2,\\
\mathbf{k}_t=(k_t^{(1)},\ldots,k_t^{(G)}),\qquad
\mathbf{c}_t=\operatorname{Concat}_{g=1}^{G}\left[\mathbf{e}_{k_t^{(g)}}^{(g)}\right].
\end{gathered}
\label{eq:grouped_quantization}
\end{equation}
Here $\mathbf{k}_t$ is the discrete index tuple, whereas $\mathbf{c}_t$ is its codebook embedding supplied to downstream decoders. Codebooks are shared across spatial positions, so the same quantization rule applies to variable-length grids at different image resolutions and aspect ratios. The resulting tuple belongs to the Cartesian product $\mathcal{K}=\{1,\ldots,K\}^{G}$, giving a combinatorial code space with
\begin{equation}
|\mathcal{K}|=K^{G},\qquad
B_{\mathrm{position}}=G\log_2 K,\qquad
B_{\mathrm{image}}=N_qG\log_2 K,
\label{eq:quantizer_capacity}
\end{equation}
where $N_q$ denotes the number of spatial positions on the quantizer grid. With $G$ codebooks of $K$ entries each, the quantizer admits up to $K^G$ combinations without explicitly storing a vocabulary of that size. At fixed $K$, increasing $G$ raises the nominal bit budget while leaving $N_q$ unchanged, but introduces more code indices to model. This budget characterizes the available discrete alphabet, not the empirical entropy or semantic information retained; greater capacity alone does not guarantee faithful reconstruction. Separate codebook lookups also do not imply statistically independent indices, as the groups encode the same visual content.

The quantization groups operate on the fused representation without predefined semantic or perceptual roles. Stacking the embeddings in spatial order gives $\mathbf{C}=[\mathbf{c}_1,\ldots,\mathbf{c}_{N_q}]^{\top}\in\mathbb{R}^{N_q\times C}$. The resulting matrix serves as a shared input to the semantic and pixel decoders, whose complementary objectives encourage the codes to retain both pretrained semantic structure and reconstruction-relevant detail. Residual modulation thus determines which information enters the bottleneck, grouped quantization supplies the capacity to encode it, and dual-decoder supervision guides how that capacity is used.

\subsection{Dual-Branch Reconstruction}
\label{sec:dual_branch}

A general-purpose visual vocabulary must preserve pretrained semantic structure while retaining the fine-grained visual detail needed to recover the original visual signal. These requirements concern different output spaces but must be satisfied
by the same discrete codes. We therefore introduce two specialized decoders that share the quantized representation.
The semantic branch aligns the codes with continuous visual features, while the pixel branch supervises image reconstruction.

\paragraph{\textbf{\upshape Representation Reconstruction Branch.}}
We take the quantized feature matrix $\mathbf{C}$, as defined in Section~\ref{sec:quantizer}, as the input to both decoders. The semantic decoder $\mathcal{D}_{\mathrm{sem}}$ maps the quantized embeddings
back to the continuous feature space after residual modulation,
\begin{equation}
    \widehat{\mathbf{Z}} = \mathcal{D}_{\mathrm{sem}}(\mathbf{C}).
\label{eq:semantic_reconstruction}
\end{equation}
We supervise this prediction with a fixed post-ResMod target
$\mathbf{Z}^{\star}$, combining feature regression with cosine alignment,
\begin{equation}
    \mathcal{L}_{\mathrm{sem}}
    = 1-\cos\!\left(\widehat{\mathbf{Z}},
    \operatorname{sg}(\mathbf{Z}^{\star})\right)
    + \lambda_{\mathrm{MSE}}^{\mathrm{sem}}
    \left\|\widehat{\mathbf{Z}}-\operatorname{sg}(\mathbf{Z}^{\star})\right\|_2^2,
\label{eq:semantic_reconstruction_loss}
\end{equation}
where $\operatorname{sg}(\cdot)$ stops gradients through the target path. The
target is produced by a frozen ResMod, as specified in
Section~\ref{sec:training_stage}. We find that cosine loss alone can be small
even for inaccurate reconstructions when a dominant shared feature mean
concentrates features in a narrow cone, consistent with representation
anisotropy~\citep{ethayarajh2019contextual}. The MSE term complements directional
alignment by constraining absolute feature values.
Reconstructing this representation encourages
the codes to retain the semantic structure inherited from the pretrained
backbone together with the information introduced by residual modulation.

\paragraph{\textbf{\upshape Pixel Reconstruction Branch.}}
Feature alignment alone does not directly constrain fidelity to the original
image. We therefore decode the same embeddings into pixels, first restoring and
expanding their spatial grid through token unmerging,
\begin{equation}
    \widehat{\mathbf{x}}
    = \mathcal{D}_{\mathrm{pix}}
    \bigl(\mathcal{U}(\mathbf{C};h_q,w_q)\bigr),
\label{eq:pixel_reconstruction}
\end{equation}
where $(h_q,w_q)$ is the quantized grid size, $\mathcal{U}$ denotes token
unmerging and spatial expansion, and $\mathcal{D}_{\mathrm{pix}}$ is the pixel
decoder. We train this branch with five complementary reconstruction losses,
\begin{equation}
    \mathcal{L}_{\mathrm{pix}}
    = \lambda_{1}\mathcal{L}_{1}
    + \lambda_{\mathrm{MSE}}^{\mathrm{pix}}\mathcal{L}_{\mathrm{MSE}}
    + \lambda_{\mathrm{LPIPS}}\mathcal{L}_{\mathrm{LPIPS}}
    + \lambda_{\mathrm{perc}}\mathcal{L}_{\mathrm{perc}}
    + \lambda_{\mathrm{Gram}}\mathcal{L}_{\mathrm{Gram}}.
\label{eq:pixel_reconstruction_loss}
\end{equation}
The $\lambda$ coefficients balance the five objectives; the pixel MSE weight
$\lambda_{\mathrm{MSE}}^{\mathrm{pix}}$ is distinct from the feature MSE weight
$\lambda_{\mathrm{MSE}}^{\mathrm{sem}}$.
All terms compare the reconstruction $\widehat{\mathbf{x}}$ with the input
$\mathbf{x}$. Here $\mathcal{L}_{1}$ and $\mathcal{L}_{\mathrm{MSE}}$ denote
mean absolute and mean squared pixel errors, respectively.
$\mathcal{L}_{\mathrm{LPIPS}}$ uses VGG features to measure perceptual similarity
across multiple layers. The ConvNeXt-S-based perceptual loss
$\mathcal{L}_{\mathrm{perc}}$ applies MSE to the network outputs to align
high-level recognition responses.
$\mathcal{L}_{\mathrm{Gram}}$ matches Gram matrices of VGG features to preserve
texture and channel correlations.
The two branches jointly constrain the information retained by the shared
vocabulary through
\begin{equation}
    \mathcal{L}_{\mathrm{dual}}
    = \lambda_{\mathrm{sem}}\mathcal{L}_{\mathrm{sem}}
    + \lambda_{\mathrm{pix}}\mathcal{L}_{\mathrm{pix}}.
\label{eq:dual_reconstruction_loss}
\end{equation}
The semantic objective anchors reconstruction in the modulated feature space,
while the pixel objective requires access to texture, color, and spatial detail. Section~\ref{sec:training_stage} describes how these objectives are introduced progressively to train the shared bottleneck and residual pathway.

\paragraph{\textbf{\upshape Flexible Encoding and Decoding at Arbitrary Resolutions.}}
We retain each image's token-grid dimensions throughout encoding and decoding. The encoder maps an input image to an $h_q\times w_q$ feature grid, which quantization converts into codebook embeddings without altering its spatial layout. For semantic reconstruction, the semantic decoder recovers a continuous feature at each grid position. For image reconstruction, $\mathcal{U}(\mathbf{C};h_q,w_q)$ uses the recorded dimensions to arrange and spatially expand the quantized tokens before the pixel decoder reconstructs an RGB image at the input resolution. Both branches therefore remain spatially aligned with the input image across varying grid dimensions and aspect ratios.

\begin{figure*}[t]
  \centering
  \includegraphics[width=\textwidth]{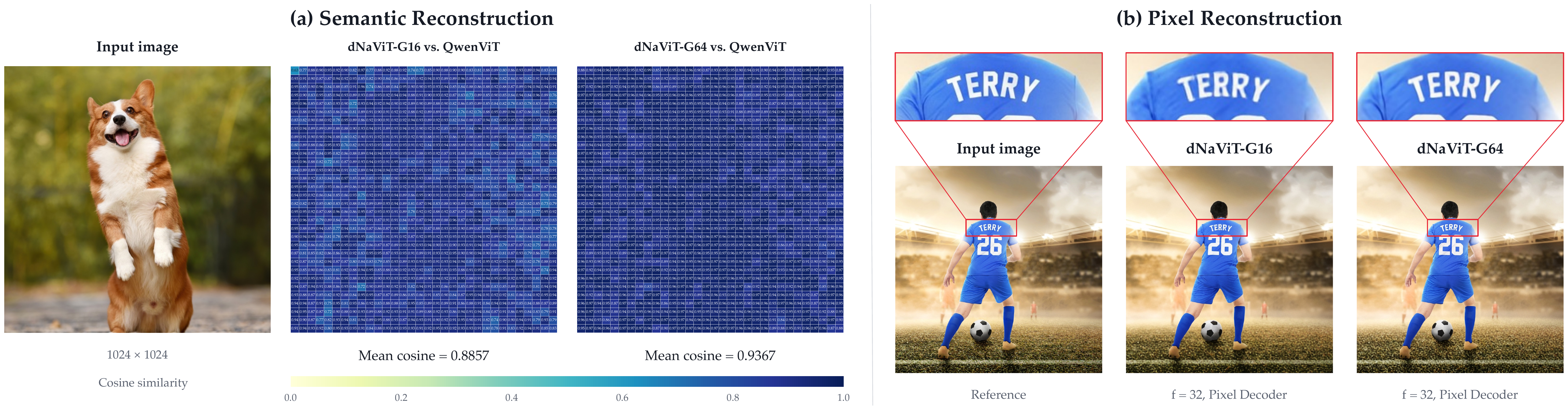}
  \caption{\textbf{Joint preservation of high-level semantics and low-level pixel detail in a shared discrete representation.}
  (a) dNaViT's reconstructed semantic
  features closely match the original QwenViT features. Each cell reports the
  cosine similarity between corresponding tokens. (b) Pixel reconstructions from dNaViT-G16 and dNaViT-G64,
  with matched crops highlighting the fine-grained text on the jersey.}
  \label{fig:dual_branch_vis}
\end{figure*}

\begin{figure*}[!t]
  \centering
  \includegraphics[width=\textwidth]{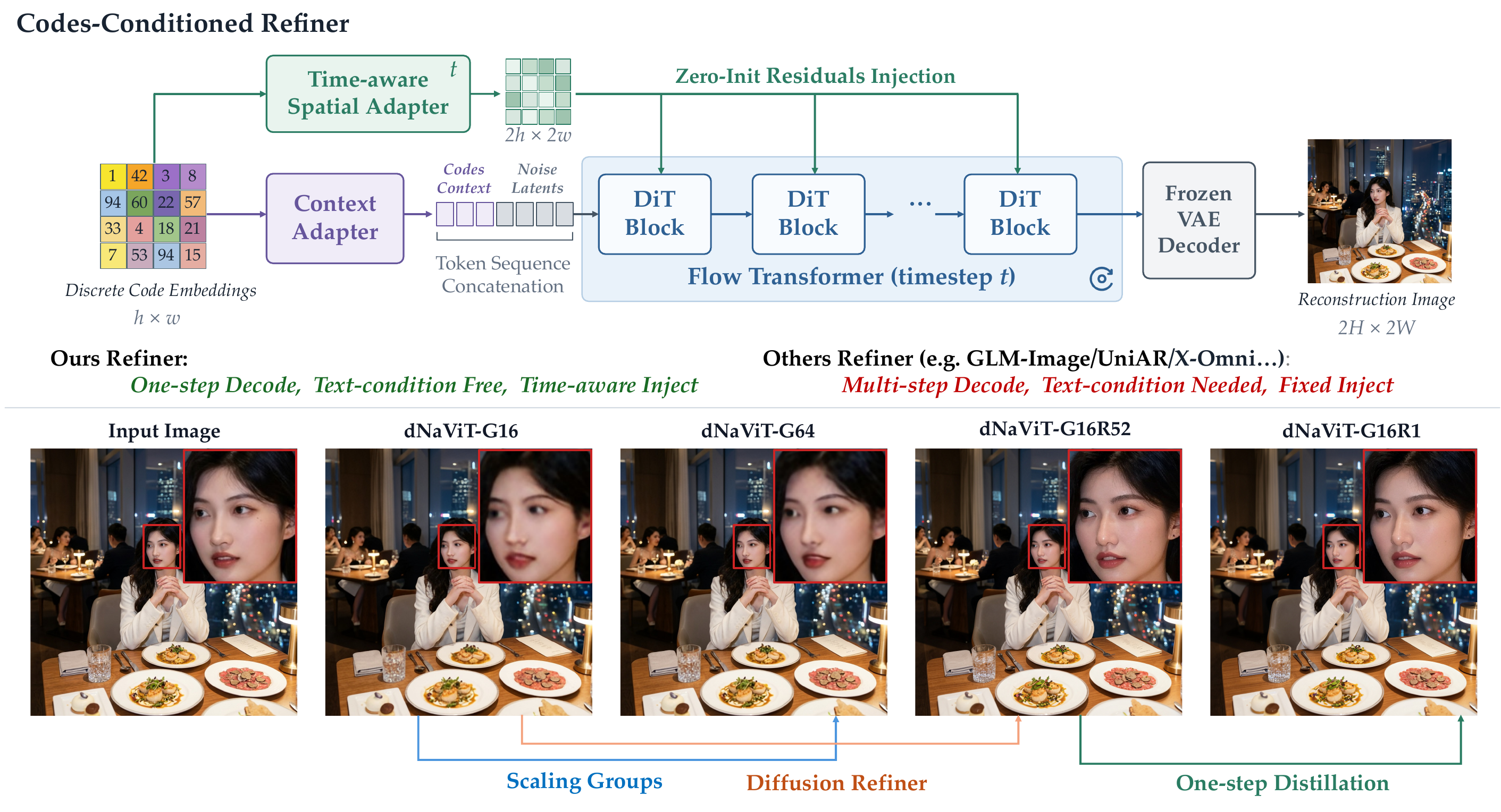}
  \caption{\textbf{Overview of the code-conditioned Refiner.}
  Code embeddings condition a flow decoder through global context tokens
  and time-conditioned spatial residuals. A multi-step teacher is distilled
  into an efficient one-step student with adversarial supervision.
  The timestep $t$ in the diagram corresponds to $\tau$ in the equations. Increasing the quantization group count (G64) improves reconstruction fidelity,
whereas combining fewer groups (G16) with the one-step Refiner preserves
fidelity while enhancing perceptual quality.}
  \label{fig:refiner_overview}
\end{figure*}

\subsection{Refiner}
\label{sec:refiner}

Although the ViT pixel decoder preserves image content faithfully, standard practices often employ a refiner module to further enhance its visual perceptual quality. We introduce a code-conditioned Refiner for
high-fidelity decoding with sharper, more realistic details
(Fig.~\ref{fig:refiner_overview}). Unlike conventional super-resolution models
conditioned on decoded RGB images, our Refiner directly conditions on discrete
visual codes, avoiding information loss introduced by intermediate pixel
decoding. It also requires no text conditioning, whose semantic priors may
alter source content and constrain image editing. Built on a pretrained
image-generation prior, the Refiner replaces the ViT pixel decoder with a
flow decoder using discrete codes as its sole conditioning input. We train
a multi-step teacher and distill it into an efficient one-step student.

\paragraph{\textbf{\upshape Code-conditioned refinement.}}
Given an encoder observation $\mathbf{x}_{\mathrm{L}}$ and a target image
$\mathbf{x}_{\mathrm{H}}$, the frozen tokenizer maps $\mathbf{x}_{\mathrm{L}}$ to codebook
embeddings $\mathbf{C}$. These embeddings condition a pretrained generative
transformer through two paths: a context adapter supplies tokens to its
joint-attention stream, and a time-conditioned spatial adapter injects
aligned residuals into the latent input and selected transformer layers.
Let $\mathbf{H}_{\mathrm{ref}}^{(j)}$ denote the token-feature matrix at
layer $j$ of the Refiner and $\tau$ denote flow time. The layer update is
\begin{equation}
\mathbf{H}_{\mathrm{ref}}^{(j+1)}
=
\mathcal{B}^{(j)}\!\left(
\mathbf{H}_{\mathrm{ref}}^{(j)},\mathcal{C}_{\phi}(\mathbf{C})
\right)
+
\mathcal{J}_{\phi}^{(j)}\!\left(
\mathcal{A}_{\phi}(\mathbf{C},\tau)
\right),
\label{eq:refiner_condition}
\end{equation}
where $\mathcal{C}_{\phi}$ forms global context tokens and
$\mathcal{J}_{\phi}^{(j)}\circ\mathcal{A}_{\phi}$ provides spatially aligned
residual features. The injection projections are zero-initialized, so the
added spatial path initially preserves the backbone computation. The global
path carries scene-level context, while the spatial path aligns generated
features with corresponding code locations. Because both adapters operate on
code token grids, the Refiner supports variable resolutions and aspect
ratios. The teacher is trained by flow matching in the latent space of a frozen
image autoencoder. Let $\mathbf{y}_0=\mathcal{V}(\mathbf{x}_{\mathrm{H}})$, and use
the noise level $\tau\in[0,1]$ as the flow coordinate, with
$\mathbf{y}_{\tau}=(1-\tau)\mathbf{y}_0+\tau\boldsymbol{\epsilon}$ and
$\boldsymbol{\epsilon}\sim\mathcal{N}(\mathbf{0},\mathbf{I})$.
We optimize
\begin{equation}
\mathcal{L}_{\mathrm{FM}}
=
\mathbb{E}\!\left[
w(\tau)\left\|
\mathbf{v}_{\theta}(\mathbf{y}_{\tau},\tau,\mathbf{C})
-
(\boldsymbol{\epsilon}-\mathbf{y}_0)
\right\|_2^2
\right].
\label{eq:refiner_flow}
\end{equation}
Here $\mathbf{v}_{\theta}$ is the conditional velocity field and $w(\tau)$
is the timestep-dependent loss weight.

\paragraph{\textbf{\upshape One-step adversarial distillation.}}
The one-step student is initialized from the frozen teacher and retains
the same code condition. Training combines latent-space
regression to the teacher's multi-step output with a stop-gradient
correction derived from the teacher's noise predictions on identically
perturbed student and teacher outputs. Adversarial supervision comes from
a time-aware discriminator  on features extracted by the frozen
teacher, using ground-truth image latents as real samples and a hinge
discriminator loss.

\section{Implementation}

\subsection{Training Stage}
\label{sec:training_stage}

\begin{figure*}[t]
  \centering
  \includegraphics[width=\textwidth]{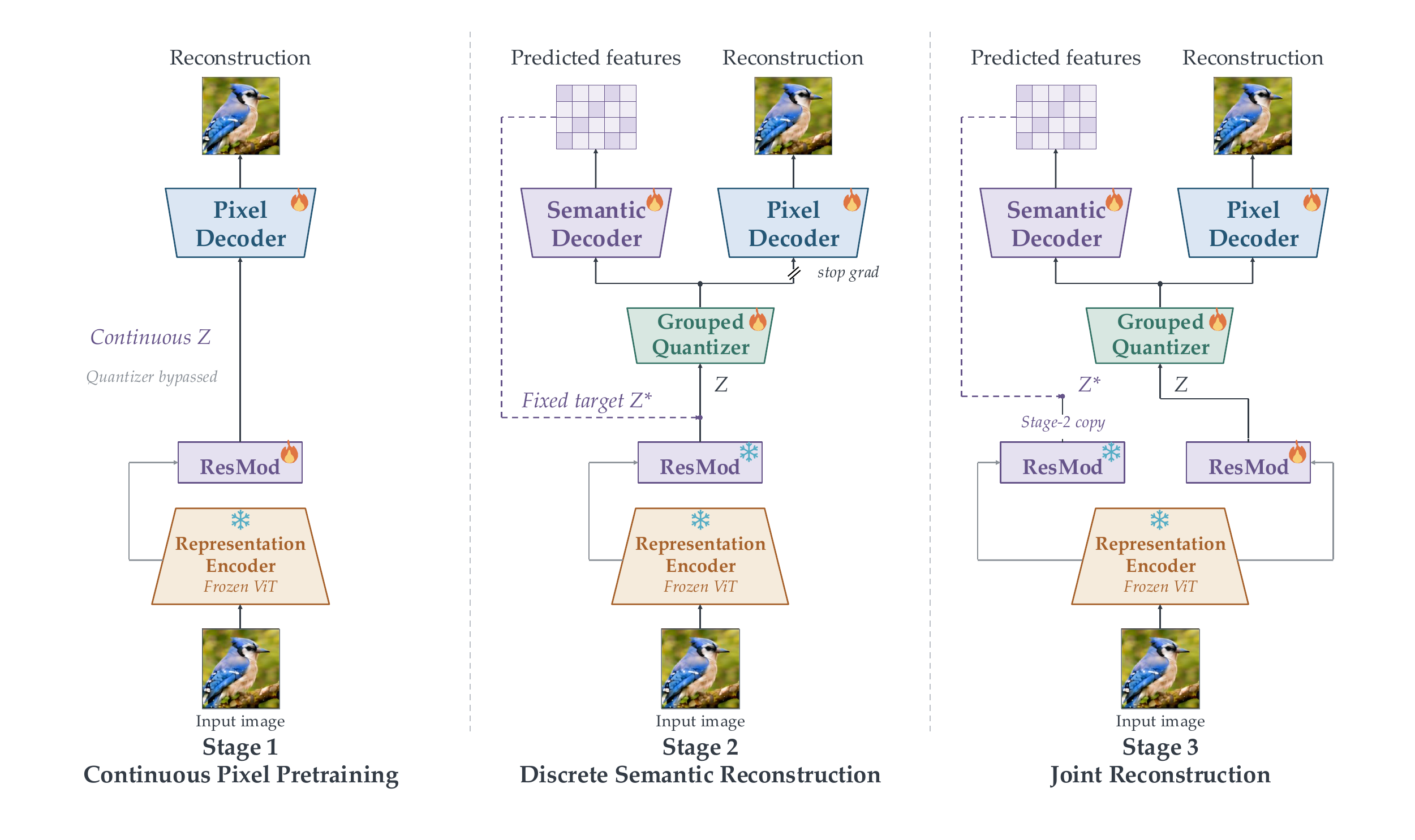}
  \caption{\textbf{Three-stage training schedule of dNaViT.}
  Continuous pixel pretraining is followed by discrete semantic reconstruction
  and joint reconstruction. Stage 2 stops pixel gradients at the quantized
  embeddings. Stage 3 opens this gradient path while a frozen copy of the
  Stage-2 ResMod supplies the feature target. Flame and snowflake icons mark
  trainable and frozen modules; dashed purple paths denote fixed targets.}
  \label{fig:training-stage}
\end{figure*}

Our optimization strategy separates learning complementary visual detail from adapting the enriched representation to a discrete bottleneck. As illustrated in Fig.~\ref{fig:training-stage}, training proceeds through continuous pixel pretraining, discrete semantic reconstruction, and joint reconstruction. The tokenizer combines a pretrained Qwen3-VL ViT ($\sim$400M), frozen throughout training to preserve its semantic representations, with a residual modulator ($\sim$100M), totaling $\sim$500M parameters. We denote by $\mathcal{E}(\mathbf{x})$ the collection of intermediate and final ViT features supplied to ResMod $\mathcal{R}$, and by $\mathcal{Q}$ the grouped quantizer.

\paragraph{\textbf{\upshape Stage 1: Continuous Pixel Pretraining.}}
We first train ResMod to recover the original visual signal from the representation encoder.
ResMod produces
$\mathbf{Z}=\mathcal{R}(\mathcal{E}(\mathbf{x}))$, which is passed
to the pixel reconstruction branch. Minimizing $\mathcal{L}_{\mathrm{pix}}$
jointly trains ResMod and the pixel decoder while the ViT
remains frozen. This stage initializes the residual pathway to expose
reconstruction-relevant information before a discrete vocabulary is
introduced.\looseness=-1

\paragraph{\textbf{\upshape Stage 2: Discrete Semantic Reconstruction.}}
This stage initializes discrete representations that preserve semantic structure
while retaining the pixel detail introduced by ResMod in Stage 1. We freeze both
the ViT and ResMod, using the ResMod output $\mathbf{Z}$ as the quantizer input and
the fixed semantic reconstruction target $\mathbf{Z}^{\star}$. The quantizer
and semantic decoder are jointly trained with $\mathcal{L}_{\mathrm{sem}}$ to
reconstruct this detail-enriched representation from
$\mathbf{C}=\mathcal{Q}(\mathbf{Z})$. In parallel, the pixel decoder is trained
with $\mathcal{L}_{\mathrm{pix}}$ on detached codes
$\operatorname{sg}(\mathbf{C})$, establishing an initial mapping from discrete
codes to pixels. Stopping gradients at this interface prevents the pixel
objective from disrupting semantic learning while the discrete
representations are being initialized.

\paragraph{\textbf{\upshape Stage 3: Joint Reconstruction.}}
This stage jointly refines the discrete representations for semantic and pixel
reconstruction. We unfreeze the encoding ResMod while retaining a frozen teacher
copy to prevent target drift during joint optimization. With student $\mathcal{R}^{\mathrm{S}}$ and
teacher $\mathcal{R}^{\mathrm{T}}$, the semantic target $\mathbf{Z}^{\star}$ now comes from the
frozen teacher, while the quantized features $\mathbf{C}$ come from the trainable student:
\begin{equation}
\begin{aligned}
    \mathbf{Z}^{\star}
    &=\mathcal{R}^{\mathrm{T}}
      \bigl(\mathcal{E}(\mathbf{x})\bigr),\\
    \mathbf{C}
    &=\mathcal{Q}\!\left(
      \mathcal{R}^{\mathrm{S}}
      \bigl(\mathcal{E}(\mathbf{x})\bigr)\right).
\end{aligned}
\end{equation}
The quantizer and both decoders continue training from Stage 2 under
$\mathcal{L}_{\mathrm{dual}}$. The semantic branch reconstructs the fixed target
$\mathbf{Z}^{\star}$, while the pixel branch is no longer detached, allowing pixel
gradients to update the quantizer and student ResMod. The ViT and teacher ResMod
remain frozen throughout.

\subsection{Training Data}
\label{sec:training_data}

The tokenization and de-tokenization training corpus combines diverse image sources to support semantic representation learning and maintain fine-grained pixel information. These sources are grouped into three complementary categories according to visual content. Tokenizer training uses images exclusively for dual-branch supervision, with a carefully controlled resolution distribution to support arbitrary-resolution training.

\paragraph{\textbf{\upshape General-Domain Images.}}
LAION~\citep{schuhmann2022laion5b}, COYO~\citep{byeon2022coyo}, and DataComp~\citep{gadre2023datacomp}, together with images from Recap-DataComp-1B~\citep{li2024recap} and CapsFusion~\citep{yu2024capsfusion}, provide broad coverage of objects, scenes, and visual styles. An in-house image corpus further expands the diversity of the training data.

\paragraph{\textbf{\upshape Text-Rich Images and Documents.}}
Selected subsets of TextAtlas5M~\citep{wang2025textatlas5m} include book covers, images containing long words, academic papers, and presentation slides. Chinese- and English-language book and news pages, along with additional OCR-oriented collections, extend coverage to dense text, small characters, and complex document layouts. Infographic collections and ChartGalaxy~\citep{li2025chartgalaxy} further contribute information-dense content, including charts and tables, where spatial relationships among visual elements are critical to interpretation.

\paragraph{\textbf{\upshape Fine-Grained and High-aesthetic Images.}}
Face-focused LAION subsets, images derived from SA-1B~\citep{kirillov2023segment}, and an in-house collection of high-aesthetic-quality images enrich the corpus with detailed facial structures, local textures, and varied image compositions.

Training begins at a fixed resolution of $256\times256$ using a limited set of image sources to facilitate rapid convergence, then progressively transitions to native-resolution training at resolutions up to 2K on a broader data mixture. During this transition, sampling maintains a balanced distribution across resolution ranges. Images are resized to approximately preserve their aspect ratios, align spatial dimensions with the encoder stride, and keep sequence lengths within the token budget. Across all stages, the model is trained on several hundred billion tokens.

\titleformat*{\paragraph}{\normalfont\bfseries}
\section{Experiment}
We evaluate dNaViT across five complementary tasks to assess the generality of its shared discrete representation. Image reconstruction measures the preservation of fine-grained detail, while frozen-feature classification and multimodal understanding assess semantic discriminability, transfer robustness, and visual reasoning. Autoregressive text-to-image generation and instruction-guided editing further examine the utility of the learned codes for conditional synthesis and image manipulation.

\subsection{Image Reconstruction}
\label{sec:image_reconstruction}

\paragraph{Evaluation Setup.}
We evaluate reconstruction on ImageNet-1K~\citep{russakovsky2015imagenet} and TokBench~\citep{wu2025tokbench} to assess whether dNaViT's shared discrete vocabulary preserves overall image fidelity and fine-grained textual and facial detail. We denote the dNaViT variants with 16 and 64 quantization groups as dNaViT-G16 and dNaViT-G64, respectively; both use a vision transformer (ViT) pixel decoder without refinement and a spatial downsampling factor of $f=32$. dNaViT-G16-R denotes the 16-group variant with the ViT pixel decoder replaced by the code-conditioned flow decoder (Refiner; Sec.~\ref{sec:refiner}). On ImageNet-1K, we report PSNR and SSIM at $256\times256$, $512\times512$, and $1024\times1024$, using the full validation set and standard preprocessing. On TokBench, we report text recognition accuracy (T-ACC), normalized edit similarity (T-NED), and face similarity (F-Sim) at $512\times512$ and $1024\times1024$, with equal weighting across the official instance-scale buckets. Baseline scores are taken from published reports. Differences in evaluation protocols and training configurations preclude controlled or rate-matched comparisons.

\begin{table}[!t]
    \centering
    \caption{\textbf{ImageNet reconstruction across resolutions.}
    $f$ denotes the spatial downsampling factor; spatial positions exclude
    quantization groups and do not specify bit rate.
    All dNaViT results use the same evaluation protocol on the full set of
    50,000 ImageNet validation images. Higher is better for both metrics.
    Baseline results follow source-specific protocols and are not directly
    comparable across sources.}
    \label{tab:imagenet_reconstruction}
    \small
    \setlength{\tabcolsep}{4pt}
    \renewcommand{\arraystretch}{1.12}
    \begin{tabular*}{\linewidth}{@{}l@{\extracolsep{0pt plus 0.75fill}}c@{\extracolsep{0pt}\hspace{9pt}}c@{\hspace{9pt}}r@{\extracolsep{0pt plus 1.25fill}}r@{\extracolsep{0pt}\hspace{24pt}}r@{}}
        \toprule
        \multirow{2}{*}{\textbf{Method}}
        & \multicolumn{3}{c}{\textbf{Configuration}}
        & \multicolumn{2}{c}{\textbf{ImageNet}} \\
        \cmidrule(lr){2-4}\cmidrule(l){5-6}
        & \textbf{Resolution} & $f$ & \textbf{\shortstack{Spatial\\positions}}
        & \textbf{PSNR (dB) $\uparrow$} & \textbf{SSIM $\uparrow$} \\
        \midrule
        \multicolumn{6}{c}{\textbf{\textit{Continuous representations}}} \\
        \midrule
        SD-VAE 1.x / 2.x~\citep{ldm} & $256^2$ & 8 & 1024 & 23.54 & 0.68 \\
        OmniTokenizer~\citep{wang2024omnitokenizer} & $256^2$ & 8 & 1024 & 26.74 & 0.82 \\
        SD-VAE XL~\citep{podell2023sdxl} & $256^2$ & 8 & 1024 & 27.37 & 0.78 \\
        Qwen-Image~\citep{wu2025qwenimage} & $256^2$ & 8 & 1024 & 32.18 & 0.90 \\
        SD-VAE 3~\citep{esser2024sd3} & $256^2$ & 8 & 1024 & 31.29 & 0.87 \\
        Wan2.1~\citep{wan2025} & $256^2$ & 8 & 1024 & 31.34 & 0.89 \\
        FLUX-VAE~\citep{flux2024} & $256^2$ & 8 & 1024 & 32.74 & 0.92 \\
        Cosmos-CI~\citep{nvidia2025cosmos} & $256^2$ & 16 & 256 & 25.07 & 0.70 \\
        VA-VAE~\citep{yao2025vavae} & $256^2$ & 16 & 256 & 27.96 & 0.79 \\
        Wan2.2~\citep{wan2025wan22} & $256^2$ & 16 & 256 & 31.25 & 0.88 \\
        PAE-DINOv2\textsuperscript{\textdaggerdbl}~\citep{yue2026pae} & $256^2$ & 16 & 256 & 27.28 & -- \\
        DC-SAE~\citep{huang2026dcsae} & $256^2$ & 32 & 64 & 23.13 & -- \\
        UniLiP~\citep{tang2025unilip} & $256^2$ & 32 & 64 & 22.99 & 0.75 \\
        UniLiP~\citep{tang2025unilip} & $448^2$ & 32 & 196 & 24.62 & 0.79 \\
        \midrule
        \multicolumn{6}{c}{\textbf{\textit{Discrete representations}}} \\
        \midrule
        ARM~\citep{wang2026arm} & $256^2$ & 16 & 256 & 19.60 & -- \\
        Cosmos-DI~\citep{nvidia2025cosmos} & $256^2$ & 16 & 256 & 19.98 & 0.54 \\
        LlamaGen~\citep{sun2024llamagen} & $256^2$ & 16 & 256 & 20.65 & 0.54 \\
        Open-MAGVIT2~\citep{luo2024openmagvit2} & $256^2$ & 16 & 256 & 22.70 & 0.64 \\
        BSQ-ViT~\citep{zhao2024image} & $256^2$ & 16 & 256 & 28.14 & 0.81 \\
        BitDance~\citep{ai2026bitdance} & $256^2$ & 16 & 256 & 24.90 & 0.72 \\
        BitDance~\citep{ai2026bitdance} & $256^2$ & 32 & 64 & 25.30 & 0.74 \\
        Emu3.5~\citep{cui2025emu35} & $512^2$ & 16 & 1024 & 23.80 & -- \\
        \dnavitrowhighlight
        \textbf{dNaViT-G16} & $256^2$ & 32 & 64 & 22.82 & 0.64 \\
        \dnavitrowhighlight
        \textbf{dNaViT-G64} & $256^2$ & 32 & 64 & 25.41 & 0.76 \\
        \dnavitrowhighlight
        \textbf{dNaViT-G16} & $512^2$ & 32 & 256 & 25.47 & 0.71 \\
        \dnavitrowhighlight
        \textbf{dNaViT-G64} & $512^2$ & 32 & 256 & 28.12 & 0.81 \\
        \dnavitrowhighlight
        \textbf{dNaViT-G16} & $1024^2$ & 32 & 1024 & 28.93 & 0.81 \\
        \dnavitrowhighlight
        \textbf{dNaViT-G64} & $1024^2$ & 32 & 1024 & 32.90 & 0.91 \\
        \bottomrule
    \end{tabular*}
    \par\vspace{4pt}
\end{table}

\paragraph{ImageNet Reconstruction.}
As shown in Table~\ref{tab:imagenet_reconstruction}, dNaViT-G64 achieves 25.41 dB PSNR with 0.76 SSIM at $256\times256$ and 28.12 dB PSNR with 0.81 SSIM at $512\times512$. At $1024\times1024$, it reaches 32.90 dB PSNR and 0.91 SSIM, compared with 28.93 dB and 0.81 for dNaViT-G16. These results indicate that dNaViT retains substantial reconstruction-relevant information within its shared discrete representation across all three resolutions.

\begin{table}[!t]
    \centering
    \caption{\textbf{Fine-grained reconstruction on TokBench.}
    Results are macro-averages over three instance-scale buckets.
    T-ACC and T-NED are percentages; F-Sim is cosine similarity. $f$ is the spatial downsampling factor.}
    \label{tab:tokbench_reconstruction}
    \small
    \setlength{\tabcolsep}{4pt}
    \renewcommand{\arraystretch}{1.12}
    \begin{tabular*}{\linewidth}{@{\extracolsep{\fill}}lcrrrrrr@{}}
        \toprule
        \multirow{2}{*}{\textbf{Method}} &
        \multirow{2}{*}{$f$} &
        \multicolumn{3}{c}{Resolution@$512\times512$} &
        \multicolumn{3}{c}{Resolution@$1024\times1024$} \\
        \cmidrule(lr){3-5}\cmidrule(l){6-8}
        & & \textbf{T-ACC$\uparrow$} & \textbf{T-NED$\uparrow$} & \textbf{F-Sim$\uparrow$}
          & \textbf{T-ACC$\uparrow$} & \textbf{T-NED$\uparrow$} & \textbf{F-Sim$\uparrow$} \\
        \midrule
        \multicolumn{8}{c}{\textbf{\textit{Continuous Autoencoders}}} \\
        \midrule
        DC-AE~\citep{chen2024dcae} & 32 & 38.25 & 56.22 & 0.39 & 52.14 & 68.02 & 0.50 \\
        VA-VAE~\citep{yao2025vavae} & 16 & 53.30 & 69.65 & 0.57 & 62.84 & 77.21 & 0.69 \\
        SD-XL~\citep{podell2023sdxl} & 8 & 56.86 & 72.55 & 0.55 & 67.84 & 81.15 & 0.70 \\
        SD-3.5~\citep{stability2024sd35} & 8 & 81.84 & 90.30 & 0.81 & 89.71 & 95.05 & 0.93 \\
        FLUX.1-dev~\citep{flux2024} & 8 & 87.64 & 93.86 & 0.86 & 93.09 & 96.96 & 0.96 \\
        \midrule
        \multicolumn{8}{c}{\textbf{\textit{Discrete Tokenizers}}} \\
        \midrule
        VQGAN~\citep{esser2021taming} & 16 & 6.12 & 17.32 & 0.19 & 15.00 & 28.80 & 0.25 \\
        Chameleon~\citep{chameleon2024mixed} & 16 & 11.55 & 26.80 & 0.28 & 23.52 & 40.39 & 0.36 \\
        LlamaGen~\citep{sun2024llamagen} & 16 & 15.01 & 30.44 & 0.25 & 27.19 & 43.58 & 0.33 \\
        VAR~\citep{tian2024var} & 16 & 29.31 & 50.00 & 0.32 & 38.36 & 59.25 & 0.40 \\
        TokenFlow~\citep{qu2025tokenflow} & 16 & 17.40 & 35.49 & 0.24 & 29.30 & 48.09 & 0.31 \\
        O-MAGVIT2~\citep{luo2024openmagvit2} & 16 & 21.65 & 39.63 & 0.31 & 34.74 & 52.03 & 0.40 \\
        O-MAGVIT2 (pretrain)~\citep{luo2024openmagvit2} & 16 & 27.33 & 47.28 & 0.31 & 39.40 & 58.95 & 0.40 \\
        UniTok~\citep{ma2025unitok} & 16 & 50.10 & 71.08 & 0.44 & 49.70 & 71.59 & 0.53 \\
        OmniTokenizer~\citep{wang2024omnitokenizer} & 8 & 42.48 & 60.70 & 0.50 & 53.48 & 70.02 & 0.61 \\
        LlamaGen~\citep{sun2024llamagen} & 8 & 52.56 & 68.99 & 0.53 & 63.89 & 78.24 & 0.67 \\
        O-MAGVIT2~\citep{luo2024openmagvit2} & 8 & 60.56 & 76.00 & 0.61 & 69.68 & 83.12 & 0.74 \\
        \midrule
        \dnavitrowhighlight
        \textbf{dNaViT-G16} & 32 & 33.58 & 52.43 & 0.34 & 45.05 & 62.75 & 0.41 \\
        \dnavitrowhighlight
        \textbf{dNaViT-G64} & 32 & 51.69 & 68.72 & 0.54 & 63.08 & 78.27 & 0.67 \\
        \bottomrule
    \end{tabular*}
    \vspace{3pt}
    \begin{minipage}{\linewidth}
    \end{minipage}
\end{table}

\paragraph{Fine-Grained Reconstruction on TokBench.}
Table~\ref{tab:tokbench_reconstruction} shows that dNaViT-G64 achieves results comparable to those reported for LlamaGen-$f8$ and higher scores than DC-AE-$f32$ on all three metrics at both resolutions. At $1024\times1024$, it attains 63.08\% T-ACC, 78.27\% T-NED, and 0.67 F-Sim. The scale-stratified results in Table~\ref{tab:tokbench_scale} further show consistent gains over dNaViT-G16 across all instance sizes. However, dNaViT-G16 and dNaViT-G64 differ in training configuration and duration, so this comparison does not isolate the effect of quantization group count. These findings indicate that dNaViT preserves textual and facial detail beyond global image structure. Small text nevertheless remains challenging, and a performance gap persists relative to stronger reconstruction models such as FLUX.1.

\begin{table}[!t]
    \centering
    \caption{\textbf{Scale-stratified reconstruction with dNaViT on TokBench.}
    S, M, and L denote the official small, medium, and large instance buckets.
    Text metrics are percentages. Buckets are resolution-specific and do not
    define a fixed-instance comparison across resolutions.}
    \label{tab:tokbench_scale}
    \small
    \setlength{\tabcolsep}{4pt}
    \renewcommand{\arraystretch}{1.12}
    \begin{tabular*}{\linewidth}{@{\extracolsep{\fill}}lcrrrrrrrrr@{}}
        \toprule
        \multirow{2}{*}{\textbf{Method}} &
        \multirow{2}{*}{\textbf{Resolution}} &
        \multicolumn{3}{c}{\textbf{T-ACC$\uparrow$}} &
        \multicolumn{3}{c}{\textbf{T-NED$\uparrow$}} &
        \multicolumn{3}{c}{\textbf{F-Sim$\uparrow$}} \\
        \cmidrule(lr){3-5}\cmidrule(lr){6-8}\cmidrule(l){9-11}
        & & \textbf{S} & \textbf{M} & \textbf{L} & \textbf{S} & \textbf{M} & \textbf{L} & \textbf{S} & \textbf{M} & \textbf{L} \\
        \midrule
        \textbf{dNaViT-G16} & $512^2$ & 3.52 & 24.56 & 72.66 & 17.60 & 52.52 & 87.19 & 0.14 & 0.25 & 0.61 \\
        \textbf{dNaViT-G64} & $512^2$ & 12.18 & 54.94 & 87.95 & 34.73 & 76.67 & 94.76 & 0.29 & 0.49 & 0.84 \\
        \midrule
        \textbf{dNaViT-G16} & $1024^2$ & 10.89 & 38.81 & 85.44 & 28.93 & 65.58 & 93.74 & 0.22 & 0.35 & 0.64 \\
        \textbf{dNaViT-G64} & $1024^2$ & 26.79 & 68.50 & 93.95 & 51.93 & 85.29 & 97.58 & 0.49 & 0.65 & 0.86 \\
        \bottomrule
    \end{tabular*}
\end{table}

\FloatBarrier

\begin{table}[!t]
    \centering
    \caption{\textbf{STEM reasoning and general multimodal understanding.}
Comprehensive comparison of vision-language models (VLMs) using continuous or discrete visual representations.}
    \label{tab:dnavit_understanding}
    \setlength{\tabcolsep}{4pt}
    \renewcommand{\arraystretch}{1.12}
    \small
    \begin{tabular*}{\linewidth}{@{\extracolsep{\fill}}lrrrrrr@{}}
        \toprule
        \textbf{Method} & \textbf{MMMU} & \textbf{MMMU-Pro} & \textbf{MathVista} & \textbf{MathVision} & \textbf{RealWorldQA} & \textbf{MMStar} \\
        \midrule
        \multicolumn{7}{c}{\textbf{\textit{Continuous}}} \\
        \midrule
        LLaVA-OV-1.5-8B~\citep{an2025llava} & 55.4 & 31.3 & 69.6 & 25.6 & 68.1 & 67.7 \\
        LLaVA-OV-2-8B~\citep{an2026llava} & -- & -- & -- & -- & 69.7 & 64.8 \\
        InternVL3-8B~\citep{zhu2025internvl3} & 62.7 & -- & 71.6 & 29.3 & 70.8 & 68.2 \\
        InternVL3.5-8B~\citep{wang2025internvl3} & 73.4 & -- & 78.4 & 56.8 & 67.5 & 69.3 \\
        Kimi-VL-A3B~\citep{kimi2025kimivl} & 57.0 & 36.0 & 68.7 & 21.4 & 68.1 & 61.3 \\
        MiMo-VL-7B-2508~\citep{xiaomi2025mimovl} & 63.1 & -- & -- & -- & -- & -- \\
        BAGEL-7B~\citep{bagel} & 55.3 & -- & 73.1 & -- & -- & -- \\
        Show-o2-7B~\citep{xie2025showo2} & 48.9 & -- & -- & -- & -- & 56.6 \\
        Molmo2-8B~\citep{clark2026molmo2} & 53.0 & -- & 58.9 & -- & 77.6 & -- \\
        Nemotron-VL-12B~\citep{nvidia2025nemotronnanov2vl} & 55.3 & -- & 69.0 & 31.5 & 76.2 & 65.9 \\
        Eagle2.5-8B~\citep{chen2025eagle25} & 55.8 & -- & 67.8 & -- & 76.7 & 66.2 \\
        Phi-4-multimodal~\citep{microsoft2025phi4mini} & 55.1 & 38.5 & 62.4 & -- & -- & -- \\
        PLM-8B~\citep{cho2025perceptionlm} & 46.1 & -- & -- & -- & 75.0 & -- \\
        Qwen2.5-VL-7B~\citep{Qwen2.5-VL} & 58.6 & 38.3 & 68.2 & 25.1 & -- & 63.9 \\
        Qwen3-VL-8B~\citep{bai2025qwen3vl} & 69.6 & 55.9 & 77.2 & 53.9 & 71.5 & 70.9 \\
        \midrule
        \multicolumn{7}{c}{\textbf{\textit{Discrete}}} \\
        \midrule
        Emu3-Chat-8B~\citep{wang2024emu3} & 31.6 & -- & 47.6\textsuperscript{*} & 15.6\textsuperscript{*} & 57.4 & 46.5\textsuperscript{*} \\
        Tar-7B~\citep{han2026vision} & 39.0 & -- & -- & -- & -- & -- \\
        ViQ-7B~\citep{yu2026viq} & 49.1 & -- & -- & -- & -- & 54.2 \\
        UniAR-8B~\citep{peng2026unified} & 44.3 & 22.7\textsuperscript{*} & 50.8\textsuperscript{*} & -- & 68.5 & 56.7\textsuperscript{*} \\
        Kelix-8B~\citep{ding2026kelix} & 54.1 & -- & 76.5 & -- & 72.1 & -- \\
        LLaDA2.0-Uni~\citep{bie2026llada2uni} & 50.1 & 34.0 & 68.1 & 26.7 & 66.7 & 64.1 \\
        \dnavitrowhighlight
        \textbf{dNaViT-8B} & 70.7 & 44.5 & 79.0 & 56.5 & 72.3 & 66.7 \\
        \bottomrule
    \end{tabular*}
\end{table}

\subsection{Image Classification}
\label{sec:image_classification}

\paragraph{Evaluation Setup.}
We evaluate frozen representations through linear probing on
ImageNet-1K~\citep{deng2009imagenet,russakovsky2015imagenet}
and direct transfer to
ImageNet-A~\citep{hendrycks2021natural},
ImageNet-R~\citep{hendrycks2021many},
ImageNet-V2~\citep{recht2019imagenet},
ImageNet-Sketch~\citep{wang2019learning}, and
ObjectNet~\citep{barbu2019objectnet}.
All representations have a $16\times16$ spatial grid before
FP32 mean pooling, followed by non-affine synchronized batch
normalization and a linear classifier.
Each encoder retains its original input resolution and normalization.
Classifiers are trained exclusively on ImageNet-1K for 10
epochs using cross-entropy and SGD with momentum 0.9, Nesterov
acceleration, zero weight decay, a global batch size of 1024,
and a peak learning rate of 0.1.
ImageNet-A and ImageNet-R use their respective 200-class subsets.
ObjectNet evaluation covers 18,574 images from the official 113
overlapping categories, with many-to-one label mapping and
red-border removal.

\begin{table}[!htbp]
    \centering
    \caption{\textbf{Image classification on ImageNet-1K and distribution-shift benchmarks.}
    Frozen representations are evaluated by linear probing on ImageNet-1K and direct transfer to ImageNet-A, ImageNet-R, ImageNet-V2, ImageNet-Sketch, and ObjectNet.
    All results report top-1 accuracy (\%).
    \textsuperscript{\dag} denotes continuous dNaViT features before quantization.}
    \label{tab:imagenet_linear_probe}
    \small
    \setlength{\tabcolsep}{4pt}
    \renewcommand{\arraystretch}{1.12}
    \begin{tabular*}{\textwidth}{@{\extracolsep{\fill}}lccrrrrrr@{}}
        \toprule
        \textbf{Method} & $f$ & \textbf{Latent Shape}
        & \textbf{IN-1K} & \textbf{IN-A} & \textbf{IN-R}
        & \textbf{IN-V2} & \textbf{IN-Sketch} & \textbf{ObjectNet} \\
        \midrule
        \multicolumn{9}{c}{\textbf{\textit{Continuous}}} \\
        \midrule
        QwenViT~\citep{bai2025qwen3vl}
        & 32 & $16\times16$
        & 80.45 & 60.13 & 65.00 & 70.18 & 46.57 & 56.48 \\
        InternViT~\citep{wang2025internvl3}
        & 28 & $16\times16$
        & 76.19 & 30.23 & 59.98 & 65.68 & 35.77 & 46.36 \\
        MoonViT~\citep{kimi2025kimivl}
        & 28 & $16\times16$
        & 75.40 & 23.96 & 49.24 & 63.34 & 34.88 & 45.46 \\
        \dnavitrowhighlight
        \textbf{dNaViT-G16}\textsuperscript{\dag}
        & 32 & $16\times16$
        & 80.40 & 60.25 & 65.19 & 70.18 & 46.56 & 56.48 \\
        \midrule
        \multicolumn{9}{c}{\textbf{\textit{Discrete}}} \\
        \midrule
        UniAR~\citep{peng2026unified}
        & 32 & $16\times16$
        & 66.54 & 14.52 & 36.13 & 54.11 & 15.40 & 29.16 \\
        UniTok~\citep{ma2025unitok}
        & 16 & $16\times16$
        & 78.56 & 47.81 & 71.10 & 68.18 & 51.03 & 60.53 \\
        TokenFlow~\citep{qu2025tokenflow}
        & 16 & $16\times16$
        & 48.32 & 6.88 & 24.81 & 37.93 & 15.35 & 15.89 \\
        Emu3.5~\citep{cui2025emu35}
        & 16 & $16\times16$
        & 8.39 & 1.16 & 3.08 & 5.87 & 0.77 & 2.71 \\
        \dnavitrowhighlight
        \textbf{dNaViT-G16}
        & 32 & $16\times16$
        & 79.47 & 56.35 & 68.09 & 69.81 & 45.02 & 59.79 \\
        \bottomrule
    \end{tabular*}
    \vspace{3pt}
\end{table}

\begin{table}[!t]
    \centering
    \caption{\textbf{OCR, document, and chart understanding.}
Comprehensive comparison of vision-language models (VLMs) using continuous or discrete visual representations.}
    \label{tab:dnavit_ocr}
    \setlength{\tabcolsep}{4pt}
    \renewcommand{\arraystretch}{1.12}
    \small
    \begin{tabular*}{\linewidth}{@{\extracolsep{\fill}}lrrrrrr@{}}
        \toprule
        \textbf{Method} & \textbf{OCRBench} & \makecell{\textbf{OCRBench V2}\\\textbf{(en/zh)}} & \makecell{\textbf{CharXiv}\\\textbf{(DQ/RQ)}} & \textbf{DocVQA} & \textbf{InfoVQA} & \makecell{\textbf{OmniDocBench}\\\textbf{(en/zh)$\downarrow$}} \\
        \midrule
        \multicolumn{7}{c}{\textbf{\textit{Continuous}}} \\
        \midrule
        LLaVA-OV-1.5-8B~\citep{an2025llava} & 829 & -- & 70.9/-- & 95.0 & 78.4 & -- \\
        LLaVA-OV-2-8B~\citep{an2026llava} & 782 & -- & -- & 95.2 & 74.4 & -- \\
        InternVL3-8B~\citep{zhu2025internvl3} & 880 & -- & 73.6/37.6 & 92.7 & 76.8 & -- \\
        InternVL3.5-8B~\citep{wang2025internvl3} & 832 & -- & 72.2/44.4 & 92.3 & 76.2 & -- \\
        Kimi-VL-A3B~\citep{kimi2025kimivl} & 867 & -- & -- & -- & 83.2 & -- \\
        MiMo-VL-7B-2508~\citep{xiaomi2025mimovl} & 881 & -- & -- & -- & -- & -- \\
        Molmo2-8B~\citep{clark2026molmo2} & -- & -- & -- & 93.2 & 80.1 & -- \\
        Nemotron-VL-12B~\citep{nvidia2025nemotronnanov2vl} & 856 & 62.0/44.2 & 76.5/41.7 & 94.7 & 79.4 & -- \\
        Eagle2.5-8B~\citep{chen2025eagle25} & 869 & -- & -- & 94.1 & 80.4 & -- \\
        Phi-4-multimodal~\citep{microsoft2025phi4mini} & 844 & -- & -- & 93.2 & 72.7 & -- \\
        PLM-8B~\citep{cho2025perceptionlm} & 870 & -- & -- & 94.6 & 80.9 & -- \\
        Qwen2.5-VL-7B~\citep{Qwen2.5-VL} & 864 & 56.3/57.2 & 73.9/42.5 & 95.7 & 82.6 & 0.31/0.40 \\
        Qwen3-VL-8B~\citep{bai2025qwen3vl} & 896 & 65.4/61.2 & 83.0/46.4 & 96.1 & 83.1 & 0.17/0.26 \\
        \midrule
        \multicolumn{7}{c}{\textbf{\textit{Discrete}}} \\
        \midrule
        Emu3-Chat-8B~\citep{wang2024emu3} & 687 & -- & 36.8\textsuperscript{*}/14.1\textsuperscript{*} & 76.3 & 42.7\textsuperscript{*} & -- \\
        X-Omni-7B~\citep{geng2025x} & 704 & -- & -- & 88.6 & -- & -- \\
        ViQ-7B~\citep{yu2026viq} & 711 & -- & -- & 88.9 & 55.3 & -- \\
        UniAR-8B~\citep{peng2026unified} & 833 & 43.0\textsuperscript{*}/48.6\textsuperscript{*} & 63.3\textsuperscript{*}/26.8\textsuperscript{*} & 91.4 & 68.9\textsuperscript{*} & 0.48\textsuperscript{*}/0.43\textsuperscript{*} \\
        Kelix-8B~\citep{ding2026kelix} & 867 & -- & -- & -- & -- & -- \\
        LLaDA2.0-Uni~\citep{bie2026llada2uni} & 757 & -- & 68.4/-- & 89.5 & 70.1 & -- \\
        \dnavitrowhighlight
        \textbf{dNaViT-8B} & 885 & 57.6/60.1 & 86.5/51.0 & 96.2 & 79.2 & 0.17/0.22 \\
        \bottomrule
    \end{tabular*}
\end{table}

\paragraph{Results.}
As shown in Table~\ref{tab:imagenet_linear_probe}, continuous dNaViT
closely matches QwenViT on ImageNet-1K and all five distribution-shift
benchmarks.
Discrete dNaViT achieves 79.47\% on ImageNet-1K, 0.93 percentage
points below its continuous counterpart.
Among the evaluated discrete representations, it achieves the highest
accuracy on ImageNet-1K, ImageNet-A, and ImageNet-V2, reaching
56.35\% and 69.81\% on the latter two.
Its transfer performance is not uniformly lower than that of
continuous dNaViT: it attains 68.09\% on ImageNet-R and 59.79\%
on ObjectNet, compared with 65.19\% and 56.48\%, respectively.
UniTok remains stronger on ImageNet-R, ImageNet-Sketch, and ObjectNet.
These results suggest that dNaViT's discrete representation preserves
substantial linearly accessible semantic information, although its
relative transfer performance depends on the target distribution.

\subsection{Visual Understanding}
\label{sec:visual_understanding}

We evaluate whether the discrete codes learned with both semantic and pixel
reconstruction retain the information needed for multimodal understanding.
We report general and mathematical reasoning separately from text, document,
and chart understanding.

\begin{table}[!t]
\centering
\small
\setlength{\tabcolsep}{4pt}
\renewcommand{\arraystretch}{1.12}
\caption{\textbf{Text-to-image generation.} GenEval2 GM is the GenEval2 overall
score; QwenImageBench is the overall score on Qwen-Image-Bench (0--100);
SpatialGenEval measures spatial-relation generation; LongText reports the
English and Chinese subsets of LongText-Bench. Comp. is per-axis spatial
compression, including latent packing. }
\label{tab:dnavit_generation}
\begin{tabular*}{\linewidth}{@{\extracolsep{\fill}}lccrrrrr@{}}
\toprule
\textbf{Model} & \textbf{Comp.} & \textbf{Params.}
& \makecell{\textbf{GenEval2}\\\textbf{GM}}
& \makecell{\textbf{QwenImage}\\\textbf{Bench}}
& \makecell{\textbf{Spatial}\\\textbf{GenEval}}
& \makecell{\textbf{LongText}\\\textbf{EN}}
& \makecell{\textbf{LongText}\\\textbf{CN}} \\
\midrule
\multicolumn{8}{c}{\textbf{\textit{Proprietary Models}}} \\
\midrule
GPT Image 1~\citep{openai2025gptimage1} & --         & --    & --    & 54.07 & 0.61  & 0.96  & 0.62  \\
Seedream 4.0~\citep{bytedance2025seedream4} & --         & --    & --    & 56.21 & 0.63  & 0.92  & 0.93  \\
Seedream 4.5~\citep{bytedance2026seedream45} & --         & --    & --    & 56.78 & --     & 0.99  & 0.99  \\
GPT Image 2~\citep{openai2026gptimage2} & --         & --    & --    & 64.69 & --     & 0.96  & 0.98  \\
Nano Banana~\citep{google2025nanobanana} & --         & --    & 44.6  & --    & 0.62  & 0.87  & 0.33  \\
Nano Banana 2.0~\citep{google2026nanobanana2} & --         & --    & --    & 59.82 & --     & 0.98 & 0.95 \\
Nano Banana Pro~\citep{google2025nanobananapro} & --         & --    & --    & 59.45 & --     & 0.98  & 0.95  \\
\midrule
\multicolumn{8}{c}{\textbf{\textit{Diffusion Models}}} \\
\midrule
HiDream-I1-Full~\citep{cai2025hidream} & $16\times$ & 17B   & --    & 43.08 & --     & 0.54  & 0.02  \\
FLUX.1 [dev]~\citep{flux2024} & $16\times$ & 12B   & 18.3    & 21.58 & 0.57  & 0.61  & 0.01  \\
BAGEL~\citep{bagel} & $16\times$ & 14B   & 23.1  & 42.04 & 0.57  & 0.37  & 0.31  \\
Z-Image~\citep{zimage2025} & $16\times$ & 6B    & 31.8    & 49.10 & --     & 0.94  & 0.94  \\
Z-Image-Turbo~\citep{zimage2025} & --         & 6B    & 31.0  & --    & 0.59  & 0.92  & 0.93  \\
Qwen-Image~\citep{wu2025qwenimage} & $16\times$ & 20B   & 33.8  & 49.23 & 0.61  & 0.94  & 0.95  \\
Qwen-Image-2512~\citep{qwen2025image2512} & --         & 20B   & 29.0  & 52.06 & 0.60  & 0.96  & 0.97  \\
OmniGen2~\citep{wu2025omnigen2} & $16\times$ & 4B    & 18.7    & 40.70 & 0.56  & 0.56  & 0.06  \\
SANA-1.5~\citep{xie2025sana15} & $32\times$ & 4.8B  & 17.5    & 35.14 & 0.54  & 0.05  & --     \\
SenseNova-U1-8B~\citep{sensenova-u1} & --    & 8B    & 40.5 & 49.02 & 0.72 & 0.95 & 0.82 \\
\midrule
\multicolumn{8}{c}{\textbf{\textit{Autoregressive Models}}} \\
\midrule
Janus-Pro-7B~\citep{chen2025januspro} & $16\times$ & 7B    & 4.6    & 23.53 & 0.51  & 0.02  & 0.01  \\
GLM-Image~\citep{zai2026glmimage} & $16\times$ & 16B   & 22.2    & 48.19 & 0.58     & 0.95  & 0.98  \\
Emu3.5~\citep{cui2025emu35} & $16\times$ & 34.1B & 25.4    & 49.89 & 0.57     & 0.98  & 0.93  \\
\dnavitrowhighlight
\textbf{dNaViT-8B} & $32\times$ & 8B & 49.5
& 52.17 & 0.68 & 0.94 & 0.94 \\
\bottomrule
\end{tabular*}
\end{table}

\paragraph{Training Setup.}
We pair the frozen visual tokenizer dNaViT-G16 with Qwen3-8B and train through alignment, pre-training, mid-training, and supervised fine-tuning using approximately 70B, 150B, 60B, and 30B tokens, respectively. Alignment uses image--text pairs, while pre-training covers image captioning, interleaved image--text sequences, OCR, grounding, multi-image data, and text-only data. Mid-training focuses on STEM and OCR, supplemented with general image--text and text-only data. Supervised fine-tuning uses visual instruction-following data covering general understanding, STEM, OCR, and document tasks.

\paragraph{Evaluation Setup.}
For general understanding, we use MMMU~\citep{yue2024mmmu},
MMMU-Pro~\citep{yue2025mmmu}, RealWorldQA~\citep{xai2024realworldqa}, and
MMStar~\citep{chen2024mmstar}; for mathematical reasoning, we use
MathVista~\citep{lu2024mathvista} and MathVision. For fine-grained visual
understanding, we evaluate OCRBench~\citep{liu2024ocrbench}, OCRBench
V2~\citep{fu2026ocrbench},
DocVQA~\citep{mathew2021docvqa}, InfoVQA~\citep{mathew2022infographicvqa},
ChartQA~\citep{masry2022chartqa}, CharXiv~\citep{wang2024charxiv}, and
OmniDocBench~\citep{ouyang2025omnidocbench}. The comparison includes models
that use continuous visual representations and models that pass visual
information through a discrete bottleneck before language-model processing.
These are model-level comparisons with potentially different training data,
image resolutions, prompting, and scoring protocols.

\paragraph{STEM and General Understanding.}
Table~\ref{tab:dnavit_understanding} demonstrates strong STEM reasoning and general understanding with discrete visual tokens. dNaViT leads the discrete models in this comparison and remains competitive with strong continuous-representation models, including comparable performance to Qwen3-VL-8B on MathVista. These results demonstrate that discrete visual representations can support strong multimodal reasoning and general understanding.

\paragraph{OCR, Document, and Chart Understanding.}
Discrete representations are often assumed to discard the fine-grained visual details required for OCR and document understanding~\citep{yu2026viq}. Table~\ref{tab:dnavit_ocr} shows that dNaViT outperforms the compared discrete models and substantially narrows the gap to strong continuous models on text-centric tasks. It performs comparably to Qwen3-VL-8B on DocVQA and OmniDocBench and leads the compared models on both CharXiv tasks. These results demonstrate that discrete visual representations can support fine-grained OCR, document understanding, and chart interpretation, challenging the view that they are inherently unsuitable for fine-grained visual tasks.

\subsection{Visual Generation and Editing}

\label{sec:visual_generation_editing}
\paragraph{Evaluation Setup.}
We evaluate text-to-image generation of dNaViT-G16-R on GenEval2,
Qwen-Image-Bench~\citep{li2026qwenimagebench}, SpatialGenEval, and
LongText-Bench~\citep{geng2025x}.
These benchmarks assess compositional alignment, prompt following,
spatial reasoning, and text rendering in English and Chinese.
For instruction-guided image editing, we use the English full set of
GEdit-Bench~\citep{liu2025step1xedit} and report semantic consistency (SC),
perceptual quality (PQ), and the benchmark's overall score.

\paragraph{Text-to-Image Generation.}
Table~\ref{tab:dnavit_generation} compares proprietary, diffusion, and
autoregressive models on compositional generation, spatial reasoning, and
text rendering, using the benchmark settings summarized in its caption.
dNaViT-8B attains the highest GenEval2 GM (49.5) among the models that
report this metric and the second-highest SpatialGenEval score (0.68),
behind SenseNova-U1-8B (0.72).
Its QwenImageBench score of 52.17 exceeds every open-weight baseline while
remaining below the proprietary systems (54.07--64.69), and it reaches
0.94 and 0.94 on the English and Chinese subsets of LongText-Bench,
within the range of the open-weight models and below the strongest
proprietary systems on Chinese text.
Baseline values are reported results and may use different training data,
sampling configurations, and evaluation protocols.

\paragraph{Instruction-Guided Image Editing.}
Table~\ref{tab:dnavit_editing} compares image editing with proprietary and
open models. SC measures consistency with the editing instruction and PQ
measures perceptual quality. We retain the benchmark-reported overall
scores rather than recomputing them as the arithmetic mean of SC and PQ.

\begin{table}[!t]
\centering
\small
\setlength{\tabcolsep}{4pt}
\renewcommand{\arraystretch}{1.12}
\caption{\textbf{Image editing on GEdit-Bench (English full set).}
SC: semantic consistency; PQ: perceptual quality. Higher is better for all
three metrics. Comp. is per-axis spatial compression, including latent
packing.}
\label{tab:dnavit_editing}
\begin{tabular*}{\linewidth}{@{\extracolsep{\fill}}lc@{\hspace{8pt}\extracolsep{0pt}}c@{\hspace{8pt}\extracolsep{\fill}}rrr@{}}
\toprule
\textbf{Model} & \textbf{Comp.} & \textbf{Params.}
& \textbf{SC $\uparrow$} & \textbf{PQ $\uparrow$} & \textbf{Overall $\uparrow$} \\
\midrule
\multicolumn{6}{c}{\textbf{\textit{Proprietary Models}}} \\
\midrule
Gemini 2.0~\citep{kampf2025gemini20image} & -- & -- & 6.73 & 6.61 & 6.32 \\
Gemini 2.5 Flash Image~\citep{fortin2025gemini25image} & -- & -- & 7.41 & 7.96 & 7.10 \\
GPT-Image-1 [High]~\citep{openai2025gptimage1} & -- & -- & 7.85 & 7.62 & 7.53 \\
\midrule
\multicolumn{6}{c}{\textbf{\textit{Open Models}}} \\
\midrule
AnyEdit~\citep{yu2024anyedit} & $8\times$ & 0.9B & 3.18 & 5.82 & 3.21 \\
Instruct-Pix2Pix~\citep{brooks2023instructpix2pix} & $8\times$ & 0.9B & 3.58 & 5.49 & 3.68 \\
MagicBrush~\citep{zhang2023magicbrush} & $8\times$ & 0.9B & 4.68 & 5.66 & 4.52 \\
UniWorld-v1~\citep{lin2025uniworld} & $16\times$ & 12B & 4.93 & 7.43 & 4.85 \\
OmniGen~\citep{xiao2024omnigen} & $16\times$ & 3.8B & 5.96 & 5.89 & 5.06 \\
FLUX.1 Kontext [Dev]~\citep{bfl2025kontext} & $16\times$ & 12B & 6.52 & 7.38 & 6.00 \\
OmniGen2~\citep{wu2025omnigen2} & $16\times$ & 4B & 7.16 & 6.77 & 6.41 \\
BAGEL~\citep{bagel} & $16\times$ & 14B & 7.36 & 6.83 & 6.52 \\
Step1X-Edit~\citep{liu2025step1xedit} & $16\times$ & 12B & 7.66 & 7.35 & 6.97 \\
Qwen-Image-Edit~\citep{qwen2025imageedit} & $16\times$ & 20B & 8.00 & 7.86 & 7.56 \\
Emu3.5~\citep{cui2025emu35} & $16\times$ & 34.1B & 8.11 & 7.70 & 7.59 \\
\dnavitrowhighlight
\textbf{dNaViT-8B} & $32\times$ & 8B & 7.41 & 7.20 & 7.30 \\
\bottomrule
\end{tabular*}
\vspace{3pt}
\end{table}

\section{Analysis}

\begin{figure}[!t]
    \centering
    \includegraphics[width=\textwidth]{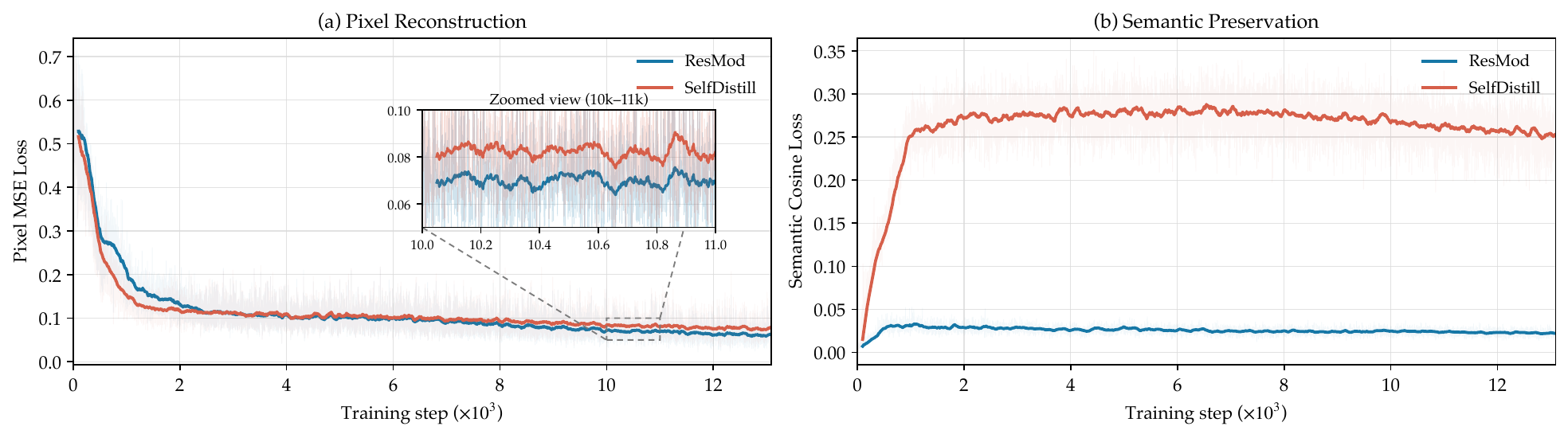}
    \caption{\textbf{Pixel and representation losses for residual feature fusion.}
    Compared with SelfDistill, ResMod maintains lower pixel MSE and representation
    cosine losses during late-stage training, indicating better preservation of
    both visual details and semantic representations.}
    \label{fig:resmod}
\end{figure}

\subsection{Effectiveness of ResMod}
\label{sec:resmod_effectiveness}
We compare ResMod (\textit{a zero-initialized residual modulation module with the pretrained encoder frozen}) with SelfDistill (\textit{a shallow fusion baseline that trains the encoder while matching its final-layer features to a frozen teacher})~\cite{unilip,uniflow}.
As shown in Fig.~\ref{fig:resmod}, ResMod maintains lower pixel MSE and representation cosine losses during the late training stage. The zoomed 10k--11k interval shows that both advantages remain stable rather than arising from isolated fluctuations. These results indicate that ResMod preserves high-level semantic representations while retaining more fine-grained visual details, resulting in more effective fusion of pixel and semantic information.

We further assess the effectiveness of ResMod on visual understanding tasks. These experiments use Qwen3-1.7B as the language backbone, with the visual tokenizer kept frozen. In Table~\ref{tab:resmod_understanding}, settings~1 and~3 train understanding models on frozen continuous features without and with pretrained ResMod, respectively. For the training-free setting~2, we directly insert pretrained ResMod into the trained model from~1 at evaluation time, without any further parameter updates. Setting~2 maintains comparable average performance to~1, even without retraining the LLM. Together with the reconstruction results above, this provides further evidence that ResMod can enrich pixel-level information while preserving pretrained semantic information.

\begin{table}[!t]
    \centering
    \caption{\textbf{Visual understanding with continuous and discrete representations.}
    Continuous features are evaluated with and without ResMod.
    OCRBench is divided by 10 when computing the
    eight-benchmark average.}
    \label{tab:resmod_understanding}
    \setlength{\tabcolsep}{4pt}
    \renewcommand{\arraystretch}{1.12}
    \small
    \resizebox{\linewidth}{!}{%
    \begin{tabular}{@{}clcccccccccc@{}}
        \toprule
        \textbf{Exp ID} & \textbf{Type} & \textbf{ResMod} & \textbf{MMMU} & \textbf{RealWorldQA} & \textbf{MMStar} & \textbf{OCRBench} & \textbf{TextVQA} & \textbf{AI2D} & \textbf{ChartQA} & \textbf{DocVQA} & \textbf{Avg. $\uparrow$} \\
        \midrule
        1 & \multirow{3}{*}{Continuous} & \ding{55} & 39.44 & 60.26 & 44.60 & 695 & 74.84 & \textbf{81.87} & \textbf{73.64} & \textbf{86.29} & 66.31 \\
        2 & & \ding{51} & \textbf{40.56} & 60.92 & 45.33 & 695 & 74.89 & 81.51 & 73.16 & 86.01 & 66.48 \\
        3 & & \ding{51} & 40.00 & \textbf{62.61} & \textbf{46.60} & \textbf{713} & \textbf{75.05} & 81.31 & 72.80 & 86.01 & \textbf{66.96} \\
        \midrule
        4 & Discrete & \ding{51} & 38.33 & 59.08 & 45.73 & 682 & 73.49 & 80.60 & 71.00 & 83.98 & 65.05 \\
        \bottomrule
    \end{tabular}}
\end{table}

\subsection{Comparison between Discrete and Continuous Approaches}
Using the same understanding training and evaluation setup as in Sec.~\ref{sec:resmod_effectiveness}, we compare discrete and continuous visual representations. Table~\ref{tab:resmod_understanding} shows comparable performance across general understanding, OCR, and document benchmarks. This observation is consistent with the scaling analysis in LongCat-Next~\citep{longcatnext2026}, which finds that discrete visual embeddings require more training data because they are learned from scratch, and that increasing the data scale progressively closes the gap with continuous models. Together, these findings suggest that discrete visual representations can match continuous features in multimodal understanding.

\subsection{Comparison of Pixel Reconstruction}

\begin{figure}[p]
    \centering
    \includegraphics[width=\textwidth]{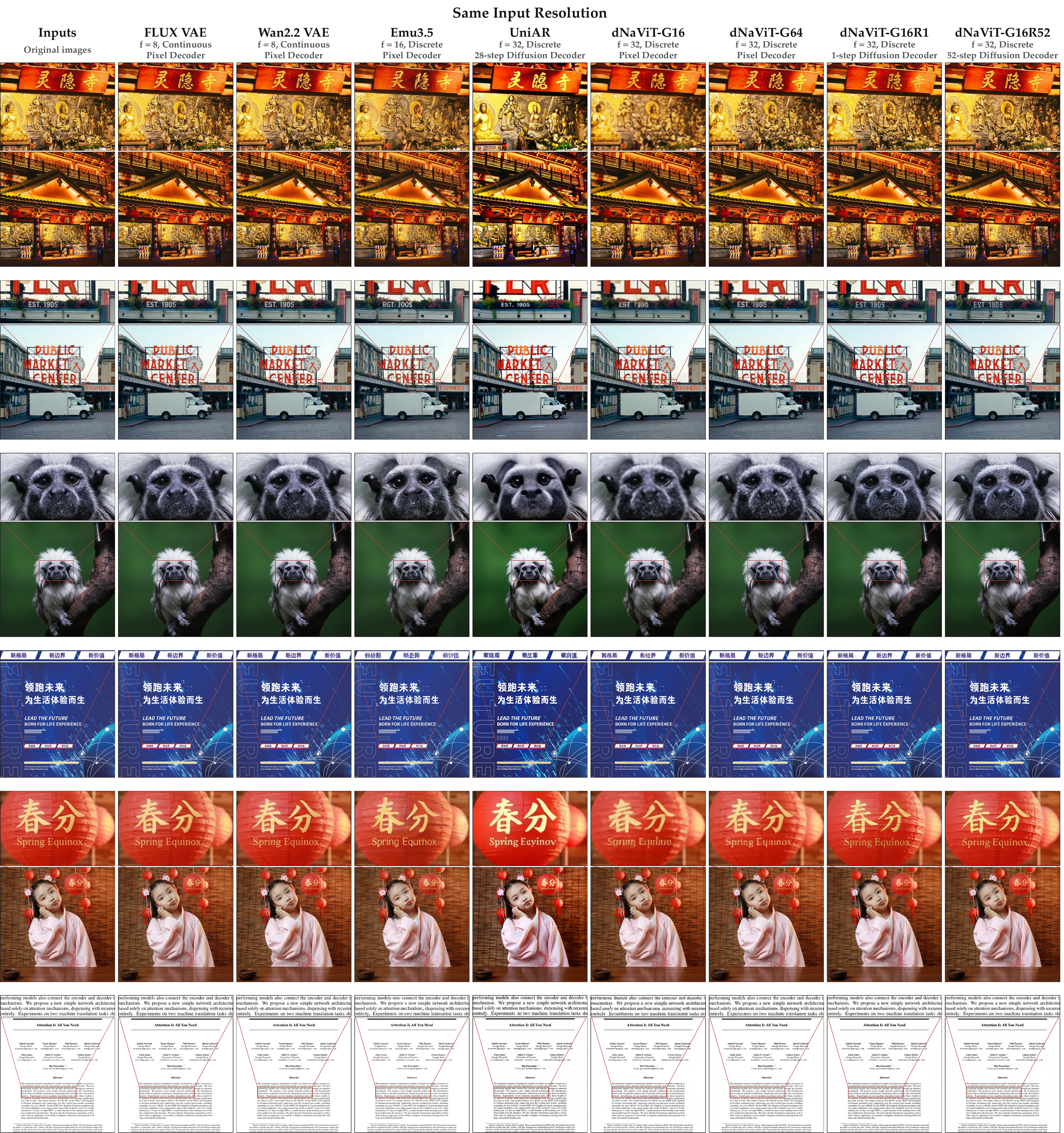}
    \caption{\textbf{Qualitative reconstruction comparison at the same input
    resolution.}
    $f$ denotes the spatial downsampling factor. Continuous and Discrete
    denote unquantized latent representations and quantized codes,
    respectively. G and R specify the number of quantization groups and
    refinement steps.}
    \label{fig:recon_compare}
\end{figure}

Fig.~\ref{fig:recon_compare} isolates the effect of the decoding strategy
while keeping the visual representation and input resolution fixed. The ViT
pixel decoder maps the discrete code embeddings directly to pixel values.
This path reliably preserves the global layout, color distribution, and
semantic content, but tends to over-smooth high-frequency information.
Consequently, small text, fine boundaries, and repetitive textures often
appear blurred or partially merged.

The flow-based Refiner, used in dNaViT-G16-R, conditions the decoder on the same
discrete representations while leveraging a pretrained generative prior.
Rather than relying solely on direct pixel regression, it recovers sharper
letter contours, finer textures, and more convincing high-frequency details.
These improvements are obtained without changing the underlying codes, and
the generated results remain consistent with the structure and semantic
content specified by the tokenizer.

The advantage becomes particularly evident under high spatial compression.
Even with a compact token representation, the Refiner continues to recover
legible text and boundary details that the pixel decoder leaves ambiguous.
This indicates that the discrete representation retains substantial
information beyond what is recovered by direct pixel decoding, and that a
generative decoder can realize this information more effectively. The
comparison therefore separates the capacity of the learned tokenizer from
the limitations of the decoding head, highlighting the benefit of our
code-conditioned refinement under highly compressed representations.

\subsection{Qualitative Visual Understanding Examples}
\label{sec:qualitative_examples}
\begin{figure}[p]
\begin{tcolorbox}[
  enhanced,
  colback=gray!2, colframe=gray!50!black,
  title=\textbf{Full-Page OCR of an 1846 Illustrated Newspaper}, fonttitle=\bfseries,
  attach boxed title to top left={yshift=-2mm,xshift=5mm},
  boxed title style={colback=gray!70!black},
  boxsep=2mm, left=2mm, right=2mm, top=5mm, bottom=2mm
]
\scriptsize
\setlength{\parskip}{0pt}
\begin{wrapfigure}{l}{0.48\linewidth}
  \vspace{-1.6\baselineskip}
  \centering
  \includegraphics[width=\linewidth]{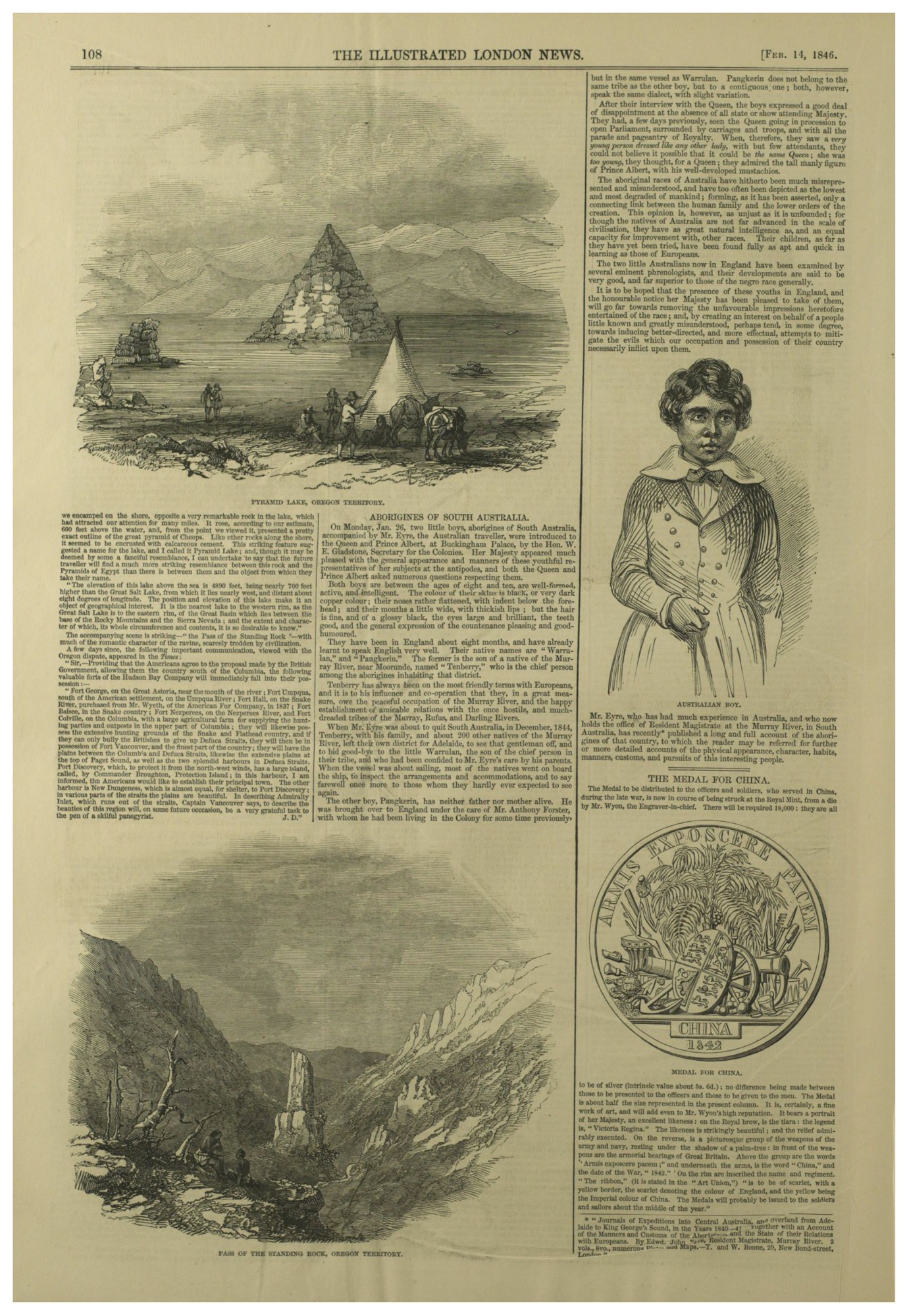}
  \vspace{-1.0\baselineskip}
\end{wrapfigure}
\raggedright
\noindent\textbf{Q:} Transcribe all text on this page verbatim in reading order, including headers, image captions, footnotes and the smallest print. Preserve the original language, paragraphs and punctuation. Do not summarize or translate.\par\vspace{0.8ex}
\textbf{A:}\par\vspace{0.3ex}
\noindent 108\par
\noindent THE ILLUSTRATED LONDON NEWS.\par
\noindent [FEB. 14, 1846.\par
\noindent PYRAMID LAKE, OREGON TERRITORY.\par
\noindent we encamped on the shore, opposite a very remarkable rock in the lake, which had attracted our attention for many miles. It rose, according to our estimate, 600 feet above the water, and, from the point we viewed it, presented a pretty exact outline of the great pyramid of Cheops. Like other rocks along the shore, it seemed to be encrusted with calcareous cement. This striking feature suggested a name for the lake, and I called it Pyramid Lake; and, though it may be deemed by some a fanciful resemblance, I can undertake to say that the future traveller will find a much more striking resemblance between this rock and the Pyramids of Egypt than there is between them and the object from which they take their name.\par
\noindent “The elevation of this lake above the sea is 4890 feet, being nearly 700 feet higher than the Great Salt Lake, from which it lies nearly west, and distant about eight degrees of longitude. The position and elevation of this lake make it an object of geographical interest. It is the nearest lake to the western rim, as the Great Salt Lake is to the eastern rim, of the great Basin which lies between the base of the Rocky Mountains and the Sierra Nevada; and the extent and character of which, its whole circumference and contents, it is so desirable to know.”\par
\noindent The accompanying scene is striking—``the Pass of the Standing Rock''—with much of the romantic character of the ravine, scarcely trodden by civilization.\par
\noindent A few days since, the following important communication, viewed on the Oregon dispute, appeared in the Times:\par
\noindent "Sir,—Providing that the Americans agree to the proposal made by the British Government, allowing them the country south of the Columbia, the following valuable forts of the Hudson Bay Company will immediately fall into their possession:—\par
\noindent ``Fort George, on the Great Astoria, near the mouth of the river; Fort Umpqua, south of the American settlement, on the Umpqua River; Fort Hall, on the Snake River, purchased from Mr. Wyeth, of the American Fur Company, in 1837; Fort Balsee, in the Snake country; Fort Nespecees, on the Nespecees River, and Fort Colville, on the Columbia, with a large agricultural farm for supplying the hunting parties and outposts in the upper part of Columbia; they will likewise possess the extensive hunting grounds of the Snake and Flathead country, and if they can only bully the British to give up Defuca Straits, they will then be in possession of Fort Vancouver; and the finest part of the country; they will have the plains between the Columba and Defuca Straits, likewise the extensive plains at the top of Paget Sound, as well as the two splendid harbours in Defuca Straits, Port Discovery, which, to protect it from the north-west winds, has a large island, called, by Commander Broughton, Protection Island; in this harbour, I am informed, the Americans would like to establish their principal town. The other harbour is New Dungeness, which is almost equal, for shelter, to Port Discovery; in various parts of the straits the plains are beautiful. In describing Admiralty Inlet, which runs out of the straits, Captain Vancouver says, to describe the beauties of this region will, on some future occasion, be a very grateful task to the pen of a skilful panegyrist. J. D.''\par
\noindent ABORIGINES OF SOUTH AUSTRALIA.\par
\noindent On Monday, Jan. 26, two little boys, aborigines of South Australia, accompanied by Mr. Eyre, the Australian traveller, were introduced to the Queen and Prince Albert, at Buckingham Palace, by the Hon. W. E. Gladstone, Secretary for the Colonies. Her Majesty appeared much pleased with the general appearance and manners of these youthful representatives of her subjects at the antipodes, and both the Queen and Prince Albert asked numerous questions respecting them.\par
\noindent Both boys are between the ages of eight and ten, are well-formed, active, and intelligent. The colour of their skins is black, or very dark copper colour; their noses rather flattened, with indent below the forehead; and their mouths a little wide, with thickish lips; but the hair is fine, and of a glossy black, the eyes large and brilliant, the teeth good, and the general expression of the countenance pleasing and good-humoured.\par
\noindent They have been in England about eight months, and have already learnt to speak English very well. Their native names are ``Warrulan,'' and ``Pangkerin.'' The former is the son of a native of the Murray River, near Moorunde, named ``Tenberry,'' who is the chief person among the aborigines inhabiting that district.\par
\noindent Tenberry has always been on the most friendly terms with Europeans, and it is to his influence and co-operation that they, in a great measure, owe the peaceful occupation of the Murray River, and the happy establishment of amicable relations with the once hostile, and much-dreaded tribes of the Murray, Rufus, and Darling Rivers.\par
\par\noindent\textit{[response continues]}
\end{tcolorbox}
\caption{Full-page OCR of a densely printed page from \textit{The Illustrated London News} (1846). The response transcribes the multi-column body text, headings and engraving captions in reading order.}
\label{fig:qual_d_en_05}
\end{figure}

\begin{figure}[p]
\begin{tcolorbox}[
  enhanced,
  colback=gray!2, colframe=gray!50!black,
  title=\textbf{Dense Chinese Academic Paper}, fonttitle=\bfseries,
  attach boxed title to top left={yshift=-2mm,xshift=5mm},
  boxed title style={colback=gray!70!black},
  boxsep=2mm, left=2mm, right=2mm, top=5mm, bottom=2mm
]
\scriptsize
\setlength{\parskip}{0pt}
\begin{wrapfigure}{l}{0.45\linewidth}
  \vspace{-1.6\baselineskip}
  \centering
  \includegraphics[width=\linewidth]{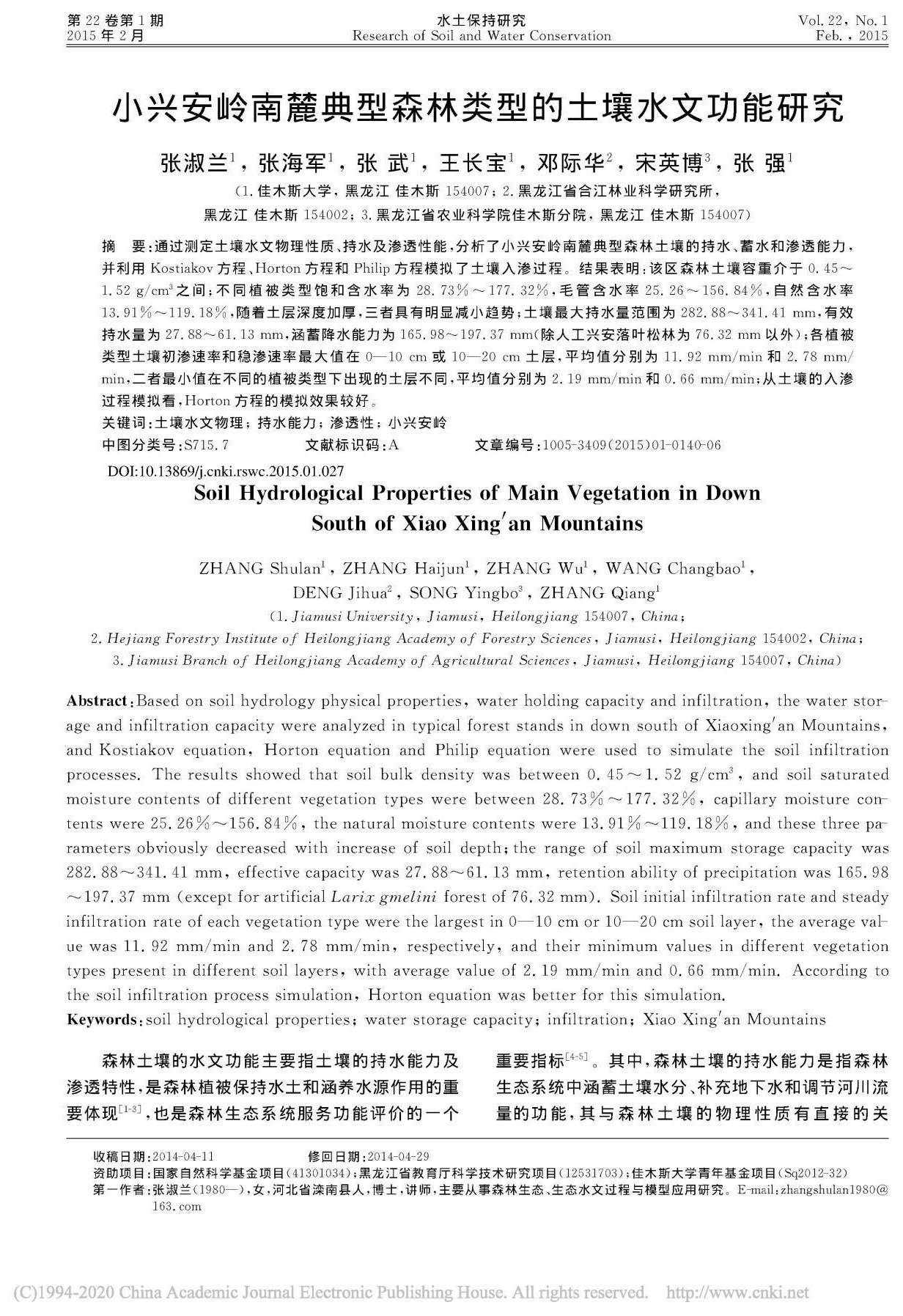}
  \vspace{-1.0\baselineskip}
\end{wrapfigure}
\raggedright
\noindent\textbf{Q:} Transcribe all text on this page verbatim in reading order, including running headers, footnotes, affiliations and the smallest print. Preserve the original language (keep Chinese text in Chinese and English text in English), paragraphs and punctuation. Do not summarize or translate.\par\vspace{0.8ex}
\textbf{A:}\par\vspace{0.3ex}
\begin{zhtext}
\noindent 第 22 卷第 1 期\par
\noindent 2015 年 2 月\par
\noindent 水土保持研究\par
\noindent Research of Soil and Water Conservation\par
\noindent Vol. 22, No. 1\par
\noindent Feb., 2015\par
\noindent\textbf{小兴安岭南麓典型森林类型的土壤水文功能研究}\par
\noindent 张淑兰 $ ^{1} $ ，张海军 $ ^{1} $ ，张武 $ ^{1} $ ，王长宝 $ ^{1} $ ，邓际华 $ ^{2} $ ，宋英博 $ ^{3} $ ，张强 $ ^{1} $\par
\noindent (1. 佳木斯大学，黑龙江 佳木斯 154002；2. 黑龙江省合江林业科学研究所，黑龙江 佳木斯 154007；3. 黑龙江省农业科学院佳木斯分院，黑龙江 佳木斯 154007)\par
\noindent 摘要：通过测定土壤水文物理性质、持水及渗透性能，分析了小兴安岭南麓典型森林土壤的持水、蓄水和渗透能力，并利用 Kostiakov 方程、Horton 方程和 Philip 方程模拟了土壤入渗过程。结果表明：该区森林土壤容重介于 0.45～1.52 g/cm $ ^{3} $  之间；不同植被类型饱和含水率为 28.73\%～177.32\%，毛管含水率 25.26～156.84\%，自然含水率 13.91\%～119.18\%，随着土层深度加厚，三者具有明显减小趋势；土壤最大持水量范围为 282.88～341.41 mm，有效持水量为 27.88～61.13 mm，涵蓄降水能力为 165.98～197.37 mm（除人工兴安落叶松林为 76.32 mm 以外）；各植被类型土壤初渗速率和稳渗速率最大值在 0—10 cm 或 10—20 cm 土层，平均值分别为 11.92 mm/min 和 2.78 mm/min，二者最小值在不同的植被类型下出现的土层不同，平均值分别为 2.19 mm/min 和 0.66 mm/min；从土壤的入渗过程模拟看，Horton 方程的模拟效果较好。\par
\noindent 关键词：土壤水文物理；持水能力；渗透性；小兴安岭\par
\noindent 中图分类号：S715.7 文献标识码：A 文章编号：1005-3409(2015)01-0140-06\par
\noindent DOI:10.13869/j.cnki.rswc.2015.01.027\par
\noindent\textbf{Soil Hydrological Properties of Main Vegetation in Down South of Xiao Xing'an Mountains}\par
\noindent ZHANG Shulan $ ^{1} $ , ZHANG Haijun $ ^{1} $ , ZHANG Wu $ ^{1} $ , WANG Changbao $ ^{1} $ , DENG Jihua $ ^{2} $ , SONG Yingbo $ ^{3} $ , ZHANG Qiang $ ^{1} $\par
\noindent 1. Jiamusi University, Jiamusi, Heilongjiang 154007, China;\par
\noindent 2. Heilongjiang Forestry Institute of Heilongjiang Academy of Forestry Sciences, Jiamusi, Heilongjiang 154002, China;\par
\noindent 3. Jiamusi Branch of Heilongjiang Academy of Agricultural Sciences, Jiamusi, Heilongjiang 154007, China)\par
\noindent Abstract:Based on soil hydrology physical properties, water holding capacity and infiltration, the water storage and infiltration capacity were analyzed in typical forest stands in down south of Xiaoxing'an Mountains, and Kostiakov equation, Horton equation and Philip equation were used to simulate the soil infiltration processes. The results showed that soil bulk density was between 0.45～1.52 g/cm $ ^{3} $ , and soil saturated moisture contents of different vegetation types were between 28.73\%～177.32\%, capillary moisture contents were 25.26\%～156.84\%, the natural moisture contents were 13.91\%～119.18\%, and these three parameters obviously decreased with increase of soil depth; the range of soil maximum storage capacity was 282.88～341.41 mm, effective capacity was 27.88～61.13 mm, retention ability of precipitation was 165.98～197.37 mm (except for artificial Larix gmelini forest of 76.32 mm). Soil initial infiltration rate and steady infiltration rate of each vegetation type were the largest in 0—10 cm or 10—20 cm soil layer, the average value was 11.92 mm/min and 2.78 mm/min, respectively, and their minimum values in different vegetation types present in different soil layers, with average value of 2.19 mm/min and 0.66 mm/min. According to the soil infiltration process simulation, Horton equation was better for this simulation.\par
\noindent Keywords: soil hydrological properties; water storage capacity; infiltration; Xiao Xing'an Mountains\par
\noindent 森林土壤的水文功能主要指土壤的持水能力及渗透特性，是森林植被保持水土和涵养水源作用的重要体现 $ ^{[1-3]} $ ，也是森林生态系统服务功能评价的一个重要指标 $ ^{[4-5]} $ 。其中，森林土壤的持水能力是指森林生态系统中涵蓄土壤水分、补充地下水和调节河川流量的功能，其与森林土壤的物理性质有直接的关\par
\noindent 收稿日期：2014-04-11 修回日期：2014-04-29\par
\noindent 资助项目：国家自然科学基金项目（41301034）；黑龙江省教育厅科学技术研究项目（12531703）；佳木斯大学青年基金项目（Sq2012-32）\par
\noindent 第一作者：张淑兰（1980—），女，河北省滦南县人，博士，讲师，主要从事森林生态、生态水文过程与模型应用研究。E-mail: zhangshulan1980@163.com\par
\noindent (C)1994-2020 China Academic Journal Electronic Publishing House. All rights reserved. http://www.cnki.net\par
\end{zhtext}
\end{tcolorbox}
\caption{Full-page OCR of a dense Chinese research article, covering the bilingual header, title, author affiliations with superscripts, abstract and body text.}
\label{fig:qual_d_zh_01}
\end{figure}

\begin{figure}[p]
\begin{tcolorbox}[
  enhanced,
  colback=gray!2, colframe=gray!50!black,
  title=\textbf{Handwritten Letter: Full Page}, fonttitle=\bfseries,
  attach boxed title to top left={yshift=-2mm,xshift=5mm},
  boxed title style={colback=gray!70!black},
  boxsep=2mm, left=2mm, right=2mm, top=5mm, bottom=2mm
]
\scriptsize
\setlength{\parskip}{0pt}
\begin{minipage}[t]{0.44\linewidth}
  \vspace{0pt}\centering
  \includegraphics[width=\linewidth,height=0.47\textheight,keepaspectratio]{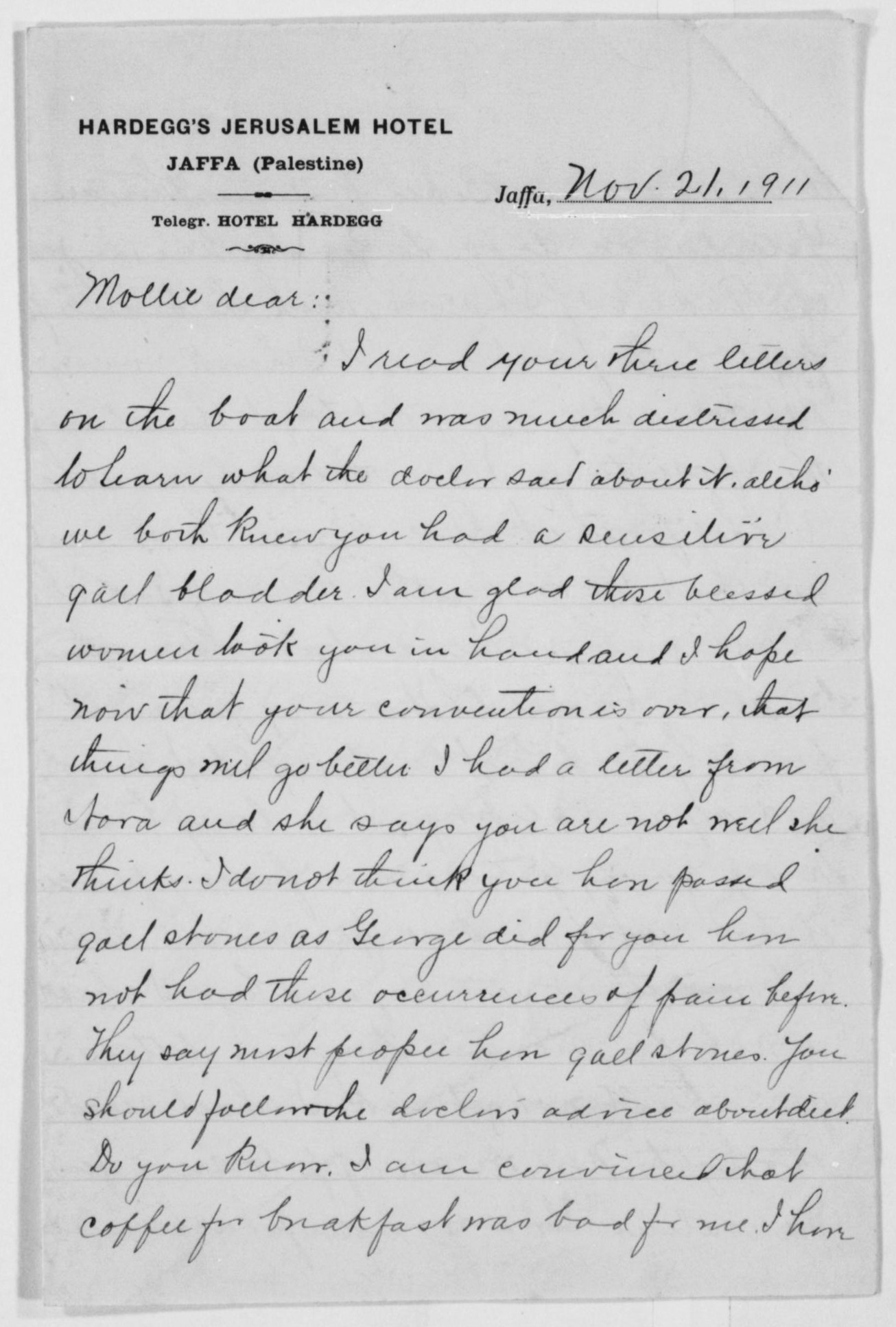}
\end{minipage}\hfill
\begin{minipage}[t]{0.53\linewidth}
\vspace{0pt}\raggedright\small
\noindent\textbf{Q:} Transcribe all handwritten text on this page verbatim, line by line, preserving the original spelling, punctuation and line breaks. Do not correct or translate. Mark illegible words as [illegible].\par\vspace{0.8ex}
\textbf{A:}\par\vspace{0.3ex}
\noindent HARDEGG'S JERUSALEM HOTEL\par
\noindent JAFFA (Palestine)\par
\noindent Telegr. HOTEL HARDEGG\par
\noindent Jaffa, Nov. 21, 1911\par
\noindent Mollie dear::\par
\noindent I read your three letters\par
\noindent on the boat and was much distressed\par
\noindent to learn what the doctor said about it, also\par
\noindent we both knew you had a sensitive\par
\noindent gall bladder. I am glad those blessed\par
\noindent women took you in hand and I hope\par
\noindent now that your convalescence is over, that\par
\noindent things will go better. I had a letter from\par
\noindent Mora and she says you are not well she\par
\noindent thinks. I do not think you have passed\par
\noindent gall stones as George did for you from\par
\noindent not had three occurrences of pain before.\par
\noindent They say most people have gall stones. You\par
\noindent should follow the doctor's advice about diet.\par
\noindent Do you know. I am convinced that\par
\noindent coffee for breakfast was bad for me. I am\par
\end{minipage}
\end{tcolorbox}
\caption{Full transcription of a 1911 cursive letter written on printed hotel stationery, covering both the printed letterhead and the connected handwriting.}
\label{fig:qual_h_en_01}
\vspace{0.6em}
\begin{tcolorbox}[
  enhanced,
  colback=gray!2, colframe=gray!50!black,
  title=\textbf{Engineering Drawing}, fonttitle=\bfseries,
  attach boxed title to top left={yshift=-2mm,xshift=5mm},
  boxed title style={colback=gray!70!black},
  boxsep=2mm, left=2mm, right=2mm, top=5mm, bottom=2mm
]
\scriptsize
\setlength{\parskip}{0pt}
\noindent{\small\textbf{Q:} Read the title block and every station range on this engineering drawing. Transcribe the drawing number, date, sheet title, project ID, sheet number and total, and list each STA.\ range under the section it belongs to. Output only the transcription.\par}\vspace{1ex}
\begin{minipage}[t]{0.50\linewidth}
  \vspace{0pt}\centering
  \includegraphics[width=\linewidth]{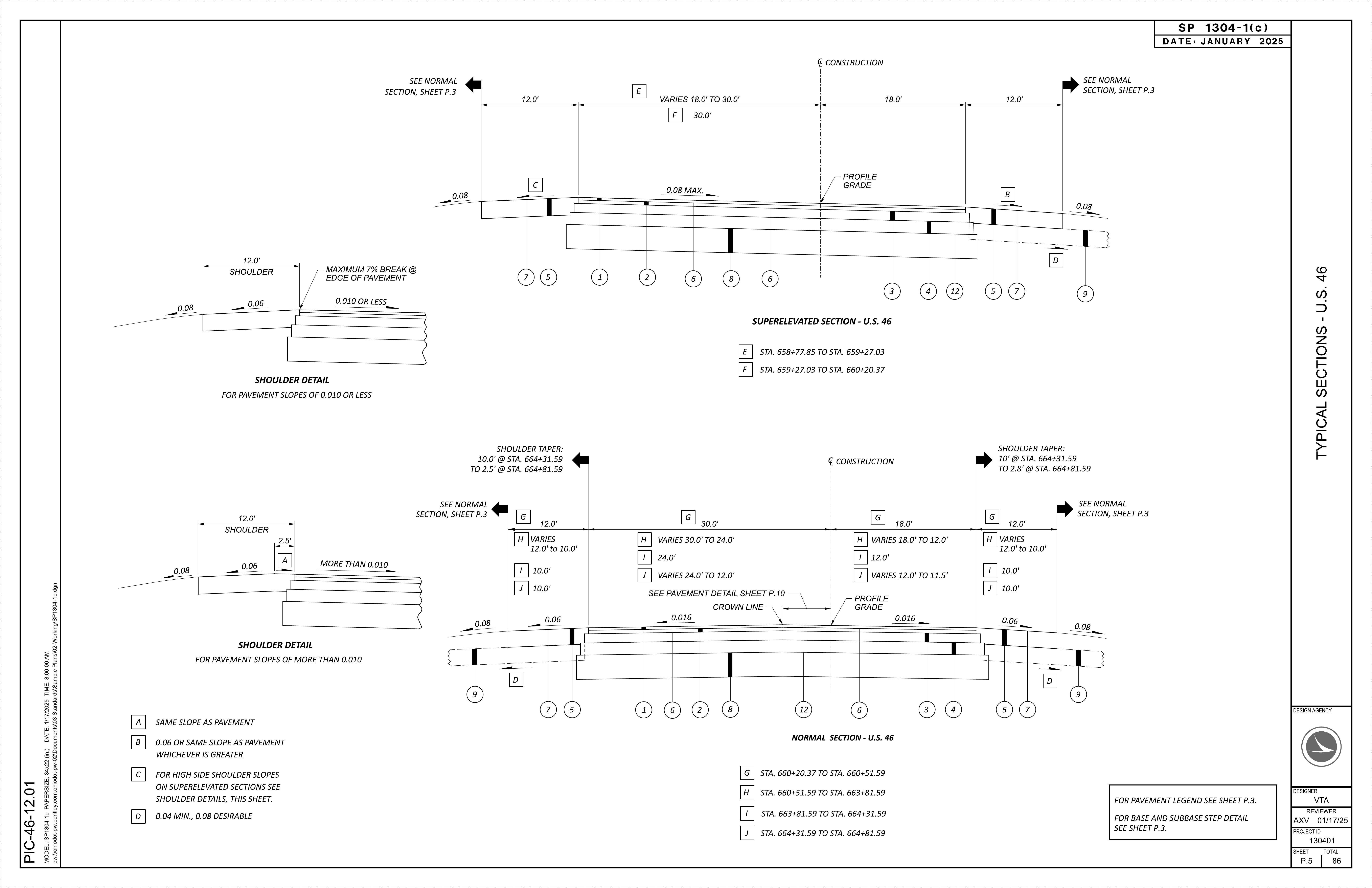}
\end{minipage}\hfill
\begin{minipage}[t]{0.47\linewidth}
\vspace{0pt}\raggedright\footnotesize
\textbf{A:}\par\vspace{0.3ex}
\textbf{Drawing Number:} SP 1304-1(c)\par
\textbf{Date:} JANUARY 2025\par
\textbf{Sheet Title:} TYPICAL SECTIONS - U.S. 46\par
\textbf{Project ID:} 130401\par
\textbf{Sheet Number / Total:} P.5 / 86\par\vspace{0.6ex}
\textbf{SUPERELEVATED SECTION - U.S. 46}\par
STA. 658+77.85 TO STA. 659+27.03\par
STA. 659+27.03 TO STA. 660+20.37\par\vspace{0.6ex}
\textbf{NORMAL SECTION - U.S. 46}\par
STA. 660+20.37 TO STA. 660+51.59\par
STA. 660+51.59 TO STA. 663+81.59\par
STA. 663+81.59 TO STA. 664+31.59\par
STA. 664+31.59 TO STA. 664+81.59\par
\end{minipage}
\end{tcolorbox}
\caption{Structured field extraction from a large-format engineering drawing: the response transcribes the title block and every station range, grouped by the typical section it belongs to.}
\label{fig:qual_w23}
\end{figure}

\begin{figure}[p]
\begin{tcolorbox}[
  enhanced,
  colback=gray!2, colframe=gray!50!black,
  title=\textbf{Equivalent Stiffness of a Spring System}, fonttitle=\bfseries,
  attach boxed title to top left={yshift=-2mm,xshift=5mm},
  boxed title style={colback=gray!70!black},
  boxsep=2mm, left=2mm, right=2mm, top=5mm, bottom=2mm
]
\scriptsize
\setlength{\parskip}{0pt}
\begin{minipage}[t]{0.30\linewidth}
  \vspace{0pt}\centering
  \includegraphics[width=\linewidth,height=0.42\textheight,keepaspectratio]{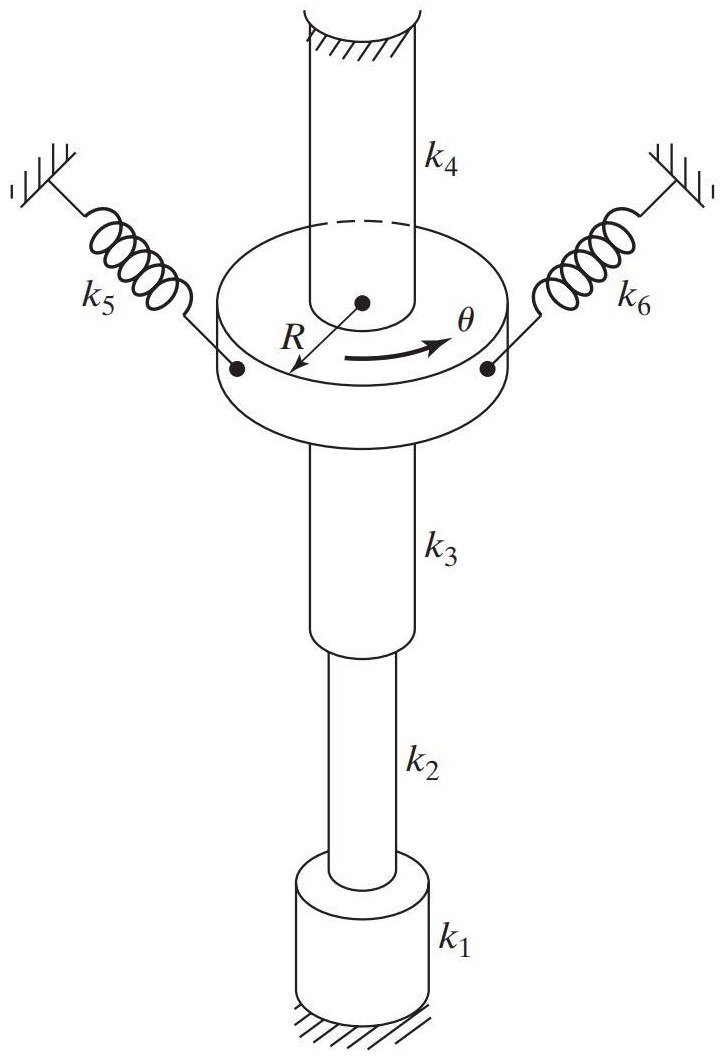}
\end{minipage}\hfill
\begin{minipage}[t]{0.67\linewidth}
\vspace{0pt}\raggedright
\noindent\textbf{Q:} Find the equivalent torsional spring constant of the system shown in the figure. Assume that k1, k2, k3, and k4 are torsional and k5 and k6 are linear spring constants. Options: A. $k_{et}=(\frac{k_{1}k_{2}k_{3}}{k_{2}k_{3}+k_{1}k_{3}+k_{1}k_{2}})+k_{5}+R^2{(k_{4}+k_{6})}$; B. $k_{et}=(\frac{k_{1}k_{2}k_{3}}{k_{2}k_{3}+k_{1}k_{3}+k_{1}k_{2}})+k_{4}+R^2{(k_{5}+k_{6})}$; C. $k_{et}=(\frac{k_{1}k_{2}k_{3}}{k_{2}k_{3}+k_{1}k_{3}+k_{1}k_{2}})+k_{6}+R^2{(k_{4}+k_{5})}$.\par\vspace{0.8ex}
\textbf{A:}\par\vspace{0.3ex}
\noindent The correct choice is \textbf{B}.\par
\par\vspace{0.3ex}
\noindent The three torsional springs \(k_1,k_2,k_3\) are in series, so their equivalent stiffness is\par
\par\vspace{0.3ex}
\noindent \[k_{123}=\frac{k_1k_2k_3}{k_1k_2+k_1k_3+k_2k_3}.\]\par
\par\vspace{0.3ex}
\noindent The linear springs \(k_5\) and \(k_6\) act at radius \(R\), so they produce torques\par
\par\vspace{0.3ex}
\noindent \[M_5=k_5R^2\theta,\qquad M_6=k_6R^2\theta,\]\par
\par\vspace{0.3ex}
\noindent giving total linear-spring torque\par
\par\vspace{0.3ex}
\noindent \[M_{56}=R^2(k_5+k_6)\theta.\]\par
\par\vspace{0.3ex}
\noindent The torsional spring \(k_4\) directly adds\par
\par\vspace{0.3ex}
\noindent \[M_4=k_4\theta.\]\par
\par\vspace{0.3ex}
\noindent Thus\par
\par\vspace{0.3ex}
\noindent \[\boxed{k_{et}=\frac{k_1k_2k_3}{k_1k_2+k_1k_3+k_2k_3}+k_4+R^2(k_5+k_6)}\]\par
\par\vspace{0.3ex}
\noindent which is \textbf{Option B}.\par
\end{minipage}
\end{tcolorbox}
\caption{Mechanical-vibration reasoning from a schematic: the model identifies series, parallel and lever-arm spring combinations (MMMU, hard).}
\label{fig:qual_e_mmmu_mechanical_engineering_29}
\vspace{1em}
\begin{tcolorbox}[
  enhanced,
  colback=gray!2, colframe=gray!50!black,
  title=\textbf{Projectile Angle at Impact}, fonttitle=\bfseries,
  attach boxed title to top left={yshift=-2mm,xshift=5mm},
  boxed title style={colback=gray!70!black},
  boxsep=2mm, left=2mm, right=2mm, top=5mm, bottom=2mm
]
\scriptsize
\setlength{\parskip}{0pt}
\begin{minipage}[t]{0.30\linewidth}
  \vspace{0pt}\centering
  \includegraphics[width=\linewidth,height=0.42\textheight,keepaspectratio]{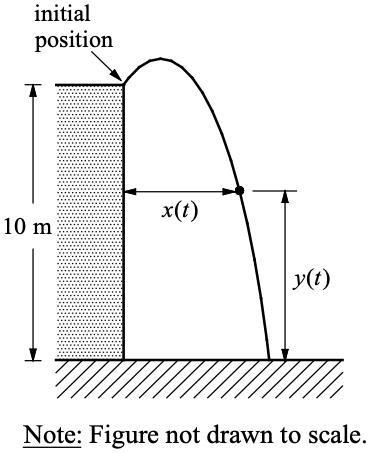}
\end{minipage}\hfill
\begin{minipage}[t]{0.67\linewidth}
\vspace{0pt}\raggedright
\noindent\textbf{Q:} An obiect is thrown upward into the air 10 meters above the ground. The figure above shows the initiaposition of the object and the position at a later time. At time t seconds after the object is thrown upwardthe horizontal distance from the initial position is given by x(t) meters, and the vertical distance from the ground is given by y(t) meters, where ${\frac{d x}{d t}}=1.4$ and ${\frac{d y}{d t}}=4.2-9.8t$,for t $\ge $ 0 .Find the angle $\theta $, 0 < $\theta $ < $\pi $/2 , between the path of the object and the ground at the instance the objecthit the ground. Options: A. $\theta $ = 0.524; B. $\theta $ = 1.047; C. $\theta $ = 1.475; D. $\theta $ = 1.570.\par\vspace{0.8ex}
\textbf{A:}\par\vspace{0.3ex}
\noindent \[y(t)=10+4.2t-4.9t^2\]\par
\par\vspace{0.3ex}
\noindent The object hits the ground when \(y(t)=0\):\par
\par\vspace{0.3ex}
\noindent \[4.9t^2-4.2t-10=0\]\par
\par\vspace{0.3ex}
\noindent \[t\approx 1.92\]\par
\par\vspace{0.3ex}
\noindent At that time,\par
\par\vspace{0.3ex}
\noindent \[\frac{dy}{dt}=4.2-9.8t\approx -14.616\]\par
\par\vspace{0.3ex}
\noindent and\par
\par\vspace{0.3ex}
\noindent \[\frac{dx}{dt}=1.4\]\par
\par\vspace{0.3ex}
\noindent So the slope of the path is\par
\par\vspace{0.3ex}
\noindent \[\frac{dy}{dx}=\frac{dy/dt}{dx/dt}\approx \frac{-14.616}{1.4}=-10.44\]\par
\par\vspace{0.3ex}
\noindent The angle with the ground is\par
\par\vspace{0.3ex}
\noindent \[\theta=\tan^{-1}(10.44)\approx 1.475\]\par
\par\vspace{0.3ex}
\noindent \textbf{Answer: C. \(\theta=1.475\)}\par
\end{minipage}
\end{tcolorbox}
\caption{Calculus reasoning grounded in a diagram: the model integrates the velocity, solves for the landing time and converts the slope into an angle (MMMU, hard).}
\label{fig:qual_e_mmmu_math_25}
\end{figure}

We present qualitative examples of our 8B understanding model on challenging visual inputs: dense full-page documents in English (Fig.~\ref{fig:qual_d_en_05}) and Chinese (Fig.~\ref{fig:qual_d_zh_01}), cursive handwriting (Fig.~\ref{fig:qual_h_en_01}), an engineering drawing (Fig.~\ref{fig:qual_w23}), and diagram-grounded reasoning (Figs.~\ref{fig:qual_e_mmmu_mechanical_engineering_29} and~\ref{fig:qual_e_mmmu_math_25}). Each example places the source image inside the model response, with the instruction and the response set. Responses are reproduced verbatim, including the model's own errors. Long transcriptions are truncated at a paragraph boundary and marked as continuing, and for examples in which the model produces an explicit reasoning trace, only the final response is shown. These examples illustrate fine-grained recognition and reasoning rather than replace quantitative evaluation.

\section{Conclusion}
\label{sec:conclude}

This work presents dNaViT, a foundational visual component acting as a native-resolution tokenizer that turns discrete codes into universal visual primitives. Through lightweight residual modulation, grouped product quantization, and progressive dual-branch optimization, dNaViT successfully reconciles high-level semantic abstraction with fine-grained visual reconstruction. Further code-conditioned refinement enhances perceptual fidelity without altering the unified representation. Across understanding, generation, and editing tasks, dNaViT matches or exceeds task-specific baselines under a significantly smaller data budget. These findings demonstrate that a shared discrete token space can serve as a universal visual currency, paving the way for fully unified, native multimodal intelligence.
\section{Contributors}
\label{sec:contributors}

Xiaotong Li\textsuperscript{*}, Zhengrong Yue\textsuperscript{*}, Xu Huang\textsuperscript{*}, Jialv Zou\textsuperscript{*}, Minhao Jing, Wenhao Yan, Haodong Li, Hai Huang, Bo Zhao, Yan Feng, Peng Pei,
Xunliang Cai

\noindent{\small\textsuperscript{*}Equal contribution.}

\clearpage

\FloatBarrier

\FloatBarrier
\bibliographystyle{assets/plainnat}
\bibliography{resources/main}
\end{document}